%% file: main.tex
\documentclass[twoside]{article}

\usepackage[accepted]{aistats2024}

\input{math_commands.tex}

\usepackage{graphicx}
\usepackage{xcolor}
\usepackage{caption}
\usepackage{subcaption}
\usepackage{amsthm}

\usepackage{enumitem}

\usepackage{booktabs}
\usepackage[linesnumbered,ruled,vlined]{algorithm2e}
\usepackage{multirow}
\usepackage{mathtools}
\usepackage{letltxmacro}
\LetLtxMacro{\originaleqref}{\eqref}
\renewcommand{\eqref}[1]{\textup{{Eq (\ref{#1})}}}

\usepackage{cleveref}
\usepackage{url}

\newcommand{\ourmethod}{MixupMP}

\newcommand{\Haug}{H^{(\text{{aug}})}}
\newcommand{\Hmixup}{H^{(\text{{Mixup}})}}
\newcommand{\FMMP}{F^{(\text{{\sc{MMP}}})}_\infty}
\newcommand{\PMMP}{\mathbb{P}^{(\text{{\sc{MMP}}})}_\infty}
\newtheorem{proposition}{Proposition}
\newtheorem{lemma}{Lemma}
\usepackage{colortbl}

\usepackage{chngcntr}

\usepackage[round]{natbib}

\newcommand{\authornote}[1]{{\let\thefootnote\relax\footnotetext{#1}}}

\begin{document}

%

%

\addtocounter{footnote}{-1}

\twocolumn[


\aistatstitle{Posterior Uncertainty Quantification in Neural Networks using Data Augmentation}

\aistatsauthor{ Luhuan Wu$^1$  \And Sinead A. Williamson }

\aistatsaddress{  Columbia University\And  Apple Machine Learning Research } ]

\addtocounter{footnote}{2}

\footnotetext{Work done during an internship at Apple Machine Learning Research.}

\addtocounter{footnote}{0}

\begin{abstract}

In this paper, we approach the problem of uncertainty quantification in deep learning through a predictive framework, which captures uncertainty in model parameters by specifying our assumptions about the predictive distribution of unseen future data. Under this view, we show that deep ensembling \citep{lakshminarayanan2017simple} is a fundamentally mis-specified model class, since it assumes that future data are  supported on existing observations only -- a situation rarely encountered in practice. To address this limitation, we propose \ourmethod{}, a method that constructs a more realistic predictive distribution using popular data augmentation techniques. \ourmethod{} operates as a drop-in replacement for deep ensembles, where each ensemble member is trained on a random simulation from this predictive distribution. Grounded in the recently-proposed framework of Martingale posteriors \citep{fong2021martingale}, \ourmethod{} returns samples from an implicitly defined Bayesian posterior. Our empirical analysis showcases that \ourmethod{} achieves superior predictive performance and uncertainty quantification on various image classification datasets, when compared with existing Bayesian and non-Bayesian approaches. 


\end{abstract}

\input{sections/intro}

\input{sections/background}

\input{sections/methods}

\input{sections/relatedworks}

\input{sections/experiments}

\input{sections/discussion}

\bibliography{refs.bib}

\input{sections/appendix}

\end{document}

%% file: math_commands.tex
\usepackage{amsmath,amsfonts,bm}

\def\eqref#1{equation~\ref{#1}}

\def\1{\bm{1}}

\DeclareMathAlphabet{\mathsfit}{\encodingdefault}{\sfdefault}{m}{sl}
\SetMathAlphabet{\mathsfit}{bold}{\encodingdefault}{\sfdefault}{bx}{n}



%% file: sections/intro.tex
\section{INTRODUCTION}\label{sec:intro}



Reliable uncertainty quantification and robust predictive performance are crucial to the deployment of deep learning models in safety-critical applications, e.g.\ medical diagnosis \citep{filos2019systematic}, autonomous driving \citep{feng2018towards} and detection of adversarial examples \citep{smith2018understanding}.  
Bayesian neural networks \citep[BNNs,][]{mackay1992practical, neal2012bayesian} have been viewed as a principled approach to achieve this goal. BNNs elicit a prior distribution over neural network parameters $\theta$. Given $n$ observations $z_{1:n}$, we estimate a posterior distribution over $\theta$, which in turn induces uncertainty in downstream predictions.
However, exact posterior inference for $\theta$ in BNNs is typically intractable, requiring approximations \citep{blundell2015weight,daxberger2021laplace}. Moreover, it can be challenging to design meaningful priors over the neural network parameter space \citep{rudner2022tractable}. 
In contrast, ``non-Bayesian" approaches like deep ensembles \citep[DE,][]{lakshminarayanan2017simple} have demonstrated promising performance in both uncertainty quantification and prediction accuracy, while maintaining simplicity in training and inference.

In this work, we  take a data-driven view of uncertainty quantification and posterior prediction for deep learning models by leveraging the assumptions on the future data distribution. 
We build on the recent idea of Martingale posteriors \citep[MPs,][]{fong2021martingale}, which transfers the prior uncertainty about parameters $\theta$ to a representation of uncertainty about future data. 
Concretely, MP specifies a \textit{predictive} distribution $\mathbb{P}_\infty(z_{n+1:\infty} | z_{1:n})$ over future data $z_{n+1:\infty}$ given the observed $z_{1:n}$. The parameters fitted on a random realization from $\mathbb{P}_\infty$ can be viewed as a posterior sample under some implicit prior \citep{doob1949application}. By fitting separate models on different random realizations, we obtain an approximate posterior distribution of parameters $\theta$. 
For example, the Bayesian bootstrap \citep[BB,][]{rubin1981bayesian} is an instance of  MP where each training dataset is drawn from a random distribution supported on existing observations. 
Unlike the explicit prior over $\theta$ required for BNNs (which can be hard to specify), practitioners can leverage domain knowledge about data to specify the predictive distribution $\mathbb{P}_\infty$.


While MP offers a fresh alternative to standard Bayesian inference, its application to deep learning has mostly remained unexplored. One reason for this is because deep learning is typically used for structured, high dimensional data, such as images or text. The predictive distributions $\mathbb{P}_\infty$ used in MP tend to make simple parametric assumptions about future observations that are inappropriate for structured data \citep{lyddon2018nonparametric,fong2019scalable},  or are not scalable to high dimensional datasets \citep[for example, the copula-based method of ][]{fong2021martingale}.



To fill in the gap between MPs and deep learning applications, we first show that the existing DE method can be viewed as an MP approach by establishing its functional equivalency to BB. However, the predictive distribution $\mathbb{P}_{\infty}$ in 
BB---and, by extension, DE--- assumes that all future data are random repetitions of the observed data $z_{1:n}$, making them fundamentally mis-specified. 
While the empirical success of BB in traditional underparametrized statistical models \citep[e.g.][]{neton1994approximate,jin2001simple} suggests that such mis-specification may be ignored in many cases, this is \textit{not} the case in a deep learning context: 
when models are over-parameterized and data is separable, BB 
is not  sufficient to represent the uncertainty in the underlying data distribution.

To address the above limitations, we propose a new predictive distribution, $\PMMP$, for deep learning models with a primary focus on the image modality. 
At a high level, $\PMMP$ introduces uncertainty in the vicinity of observations, which increases when future data is more dissimilar from observations. 
To achieve this goal, we take inspiration from the Dirichlet process, which
combines the empirical distribution of $z_{1:n}$ with samples from some base measure $H$. We replace the empirical component of Dirichlet process (which assumes exact repeats of observations $z_{1:n}$) with a version that allows for augmented repeats of $z_{1:n}$, thereby adding plausible samples that are similar to each observation. We then specify the base measure in terms of Mixup \citep{zhang2017mixup}, a data augmentation technique that linearly interpolates random pairs of images and their labels. Samples from $\PMMP$ yield a diverse set of future observations with low label uncertainty near our observations $z_{1:n}$, and higher label uncertainty as we move further from $z_{1:n}$. Such behavior is aligned with previous work that suggests that it often suffices to impose uncertainty on the boundary of the training data, rather than across the entire input space \citep{lee2017training, hafner2020noise}.


We show that $\PMMP$ can be used in a simple, ensemble-like procedure which we call \ourmethod{}, by  training multiple models  on different future data streams drawn from $\PMMP$.  Grounded by the MP framework, the fitted parameters of each model can be viewed as a valid posterior sample for $\theta$. 
Additionally, we  devise an efficient, single-model variant of \ourmethod{} that leverages implicit ensemble techniques introduced by \citet{gal2016dropout}.

To summarize our contributions, (1) we demonstrate that, in a deep learning context, DE is equivalent to BB, and therefore is a form of MP; however, we argue that this form of MP is mis-specified for deep learning applications; (2) we develop \ourmethod{}, a novel MP formulation suitable for deep learning using image data; (3) we show, through empirical study, that \ourmethod{} can outperform existing ensemble-based approaches and other approximate Bayesian methods in predictive performance and uncertainty calibration. 

%% file: sections/background.tex
\section{Background}\label{sec:background}
\paragraph{Setup and notation.} We focus on the supervised learning setting where we have i.i.d. samples $\{z_i=(x_i, y_i)\}_{i=1}^n$, with $x_i$  the input and $y_i$ the class label belonging to one of the $K$ classes. A model then learns a parameterized distribution $p_\theta (y=k | x)$ for $k=1:K$ by optimizing some loss function $l_\theta (\cdot)$ over the data. 

\subsection{Martingale posterior distributions}\label{sec:bg_mp}

The martingale posterior \citep[MP,][]{fong2021martingale} is a recently proposed uncertainty quantification technique that offers an alternative to classical Bayesian inference.  Rather than specify a prior 
over parameters, they posit a distribution $\mathbb{P}_\infty$ over unseen future observations $z_{n+1:\infty}$, given the observed data $z_{1:n}$.
The martingale posterior is then defined as
\begin{align}\label{eq:mp}
    \textstyle \pi_n(\theta) = \int \theta^*(z_{1:\infty}) d\mathbb{P}_{\infty}(z_{n+1:\infty}|z_{1:n})
\end{align}
where $\theta^*(z_{1:\infty})$ is the estimator of the parameters of interest $\theta$ given $z_{1:\infty}$, typically obtained by minimizing the loss  $\ell_\theta (z_{1:\infty})$.  If $\mathbb{P}_\infty$ is a martingale, then  $\pi_n(\theta)$  converges to the Dirac measure centered at the true $\theta$ as $n\rightarrow \infty$ (up to an equivalency set). Furthermore, $\pi_n (\theta)$ can be seen as a Bayesian posterior under some (typically unknown) prior on $\theta$, a result of Doob's Theorem \citep{doob1949application,fong2021martingale}. 

The martingale requirement is satisfied if the sequence $z_{n+1:\infty}$  is conditionally identically distributed \citep{berti2004limit}. A weaker requirement, that we use in this paper, is that the sequence be infinitely exchangeable --- i.e., $z_i\stackrel{\small{iid}}{\sim}F_\infty$ for $i>n$ given some latent measure $F_\infty$ \citep{de1937prevision}. 
One convenient construction for $F_\infty$ is the posterior of a Dirichlet process (DP) with base measure $H$ and concentration parameter $c$ \citep[as proposed by][]{fong2019scalable},
\begin{align}
\label{eq:dp}
    \textstyle F_\infty 
    \sim \mbox{DP}\left(c+n, \frac{cH + \sum_{i=1}^n \delta_{z_i}}{c+n}\right),
\end{align}
where $F_\infty:= \sum_{i=1}^n w_i \delta_{z_i} + \sum_{i=n+1}^\infty w_i \delta_{\phi_i}$, with $\phi_i \sim H$. For the first $n$ weights (corresponding to the original  $z_{1:n}$), we have $\mathbb{E}[w_i]=1/(n+c)$; for the tail we have $\mathbb{E}\left[\sum_{j=n+1}^\infty w_j\right] = c/(n+c)$.
This formulation captures the idea that future data are likely to be a combination of repeats of  empirical observations $z_{1:n}$, and samples from a distribution $H$ that captures  prior beliefs or some data-driven centering model, with $r=c/n$ capturing the ratio of the two.




A sample $F_\infty^{(b)} = \sum_{i=1}^n w_i^{(b)}\delta_{z_i} + \sum_{i=n+1}^\infty w_i^{(b)}\delta_{\phi_i^{(b)}}$ from \Cref{eq:dp} describes the  empirical distribution of a sampled sequence $z_{1:\infty}^{(b)}$\footnote{Since $n$ is finite, the distribution of $z_{1:\infty}$ is almost surely equal to the distribution of $z_{n+1:\infty}$.}. One can further obtain a sample  $\theta^{(b)}$ from the corresponding martingale posterior as $\theta^{(b)} =\arg \min_\theta  \int \ell_\theta (z_{1:\infty}) dF_\infty^{(b)}(z_{1:\infty}) = \arg\min_\theta \sum_{i=1}^n w_i^{(b)} \delta_{z_i} + \sum_{i=n+1}^\infty w_i^{(b)} \delta_{\phi_i^{(b)}}$.
In practice, we can approximate the infinite $F_\infty^{(b)}$ with a finite measure. 
The full procedure to  to generate $B$ posterior samples from the DP-based martingale posterior (DP-MP) is given in \Cref{alg:dp-mp}. 

\begin{algorithm}[t]
\caption{Dirichlet Process-based Martingale Posterior (DP-MP)}
\label{alg:dp-mp}
\KwIn{Observation $z_{1:n}$; DP base measure $H$ and concentration parameter $c$, approximation value $T$, \# of simulations $B$, loss  $l_\theta(\cdot)$}
\KwResult{Posterior samples $\{\theta^{(b)} \}_{b=1}^B$}

\For{$b = 1$ \KwTo $B$}{
    Draw pseudo data $z^{(b)}_{n+1:n+T} \sim H$ \
    
    Draw weights 
    \begin{equation*}w_{1:n+T}^{(b)} \sim \mbox{Dirichlet}(\underbrace{1,\cdots,1}_{\text{length }n}, \underbrace{c/T, \cdots, c/T}_{\text{length } T})\label{eqn:approxDP}\end{equation*}\
    
    Compute $\theta^{(b)} = \arg\min_\theta \sum_{i=1}^{n+T} w_i l_\theta (z_i)$
}
\end{algorithm}



\paragraph{Bayesian bootstrap.} 
If $c=0$ in the above construction, we recover the Bayesian bootstrap \citep[BB,][]{rubin1981bayesian}, where $F_\infty^{(b)}=\sum_{i=1}^n w_i^{(b)}\delta_{z_i}$, with
\begin{align}
        (w_1^{(b)},\dots, w_n^{(b)}) \sim& \mbox{Dirichlet}(1,\dots, 1).
    \end{align}
We then optimize the weighted loss to obtain 
\begin{align}
    \theta^{(b)} &\textstyle = \arg \min_\theta \sum_{i=1}^n w_i^{(b)} l_\theta (z_i).\label{eqn:weighted_erm}
\end{align}
BB is a special case of the MP that assumes all future observations are repeats of the $n$ original observations. 

\subsection{Ensemble methods in deep learning}
\paragraph{Deep ensembles.} A deep ensemble \citep[DE,][]{lakshminarayanan2017simple} is a collection of $B$ neural networks, typically with the same architecture, trained from different random initializations of $\theta$. Specifically, each network obtains its set of parameters $\{\theta^{(b)}\}_{b=1}^B$ by minimizing the  empirical loss, 
\begin{align}
    \theta^{(b)} &= \textstyle \arg \min_\theta \sum_{i=1}^n l_\theta (z_i) \label{eq:erm}.
\end{align}

The prediction is then made by combining outputs from individual networks, e.g. averaging their classification logits. 
DE has been shown to perform similarly to or better than alternative Bayesian neural networks in uncertainty quantification, predictive accuracy and robustness to distribution shifts \citep{ovadia2019can}.


\paragraph{Monte Carlo dropout.} Monte Carlo dropout \citep[MC Dropout,][]{gal2016dropout} trains a single neural network using the dropout technique \citep{srivastava2014dropout}. During inference, MC Dropout simulates an ensemble of models by randomly activating dropout in separate forward passes. This procedure can be interpreted either as an approximation to DE \citep{hara2016analysis} or as a variational inference method for BNN \citep{gal2016dropout}.





\subsection{Mixup}
Mixup \citep{zhang2017mixup} is a data augmentation technique that samples an augmentation $z'=(x', y')$ as a convex combination of two random observations $z_i=(x_i, y_i)$ and $z_j=(x_j, y_j)$ from $z_{1:n}$,\footnote{In this paper, we assume a separate $\lambda$ is generated for each sample $z'\sim \text{Mixup}(z_{1:n};\alpha)$.}
\begin{equation}
\begin{rcases}
\lambda &\sim \mbox{Beta}(\alpha, \alpha)\\
    z_i, z_j &\stackrel{\small{iid}}{\sim} \{z_1,\dots, z_n\}\\
    x' &= \lambda x_i + (1-\lambda)x_j\\
    y' &= \lambda y_i + (1-\lambda)y_j\\
    \end{rcases}
    \quad z' \sim \text{Mixup}(z_{1:n}; \alpha)\label{eqn:mixup}
\end{equation}
for some $\alpha>0$. 
Directly training a model on these augmented examples has been found to improve test set accuracy and calibration performance \citep{zhang2017mixup,thulasidasan2019mixup}. When used in an ensemble, however, Mixup has been shown to have worse uncertainty calibration performance than the single-model Mixup  or deep ensemble trained on original data \citep{rahaman2021uncertainty,wen2020combining}.

%% file: sections/methods.tex
\section{An equivalency between DE and BB}\label{sec:equivalency} 

\input{sections/theory}

\section{Mixup Martingale posteriors: Incorporating prior knowledge about the distribution of interest}\label{sec:mixupmp}

Instead of concentrating on  existing observations as in BB or DE, a better choice of the predictive distribution would allow for plausible additional observations that reflect our beliefs and uncertainty about future data. The DP-MP approach discussed in \Cref{sec:background} 
adds in such additional observations. For low-dimensional or unstructured data we can obtain plausible new samples using relatively simple base measures (see \Cref{app:sec:synthetic} for examples). However,  such measures do not translate well to highly structured data such as images: sampling from a moment-matched Gaussian on pixel space will not yield realistic images, for example.

Instead, we propose \ourmethod{}, an MP  method that models the predictive uncertainty with  data augmentation techniques. We focus on image data, for which data augmentation has proved highly successful \citep{shorten2019survey}, although the general ideas could be extended to alternative modalities \citep[e.g.,][]{feng2021survey,meng2021mixspeech}.

We start from the DP-MP approach, which draws samples of future data that are either copies of previous observations, or ``new'' observations from a base measure $H$. In practice we do not expect future data to be exact copies of $z_{1:n}$; instead we assume  variations of original data obtained by standard randomized label-preserving data augmentation techniques such as random cropping.  Hence we replace the point masses $\delta_{z_i}$ in \Cref{eq:dp} with a distribution $\Haug_{z_i}(\cdot)$, such that samples from $\Haug_{z_i}$ are generated by (i) sampling a random augmentation from some set $\mathcal{T}^{\text{aug}}$, (ii) applying that augmentation to $x_i$, and (iii) combining with the original $y_i$. 

\begin{algorithm}[!t]
\caption{\ourmethod{} using mini-batches} 
\label{alg:ourmethod}
\KwData{Data $\{z_i=(x_i,y_i)\}_{i=1}^n$, concentration ratio $r:=c/n$, Mixup parameter $\alpha$, minibatch size $n_{mb}$, pseudosample batch size $t_{mb}$, \# of simulations $B$, loss $l_\theta(\cdot)$, data loader $d_{n_{mb}}(z_{1:n})$}
\KwResult{Posterior samples $\{\theta^{(b)} \}_{b=1}^B$}  
\For{$b = 1$ \KwTo $B$}{

    Randomly initialize $\theta^{(b)}$

    \While{not converged}{
    \For{$z^{(mb)}_{1:n_{mb}}$ \textbf{in} $d_{n_{mb}}(z_{1:n})$}{
    \label{line:random-sample-mini-batch}
    \For{$i=1$ \KwTo $n_{mb}$}{
    Sample $\tilde{z}_i \sim \Haug_{z^{(mb)}_i}$ \
    }
    \For{$i=n_{mb}+1$ \KwTo $n_{mb} + t_{mb}$}{
    Sample $\tilde{z}_i \sim \Hmixup_{\tilde{z}_{1:n_{mb};\alpha}}$\label{line:mixup-sample}
    } 


    Update $\theta^{(b)}$ by  gradient descent on 
     $\sum_{i=1}^{n_{mb}} \ell_{\theta^{(b)}} (\tilde z_i) + r\frac{n_{mb}}{t_{mb}} \sum_{i=n_{mb} + 1}^{n_{mb}+t_{mb}} \ell_{\theta ^{(b)}} (\tilde {z}_{i})$    
    }
    }
}
\end{algorithm}

Second, we specify a data-driven base measure $\Hmixup_{z_{1:n}; \alpha}$. We sample from $\Hmixup_{z_{1:n}; \alpha}$ by first applying random augmentations to $x_{1:n}$ to get augmented inputs $\tilde{x}_{1:n}$, then sampling an observation $z'$ according to Mixup($\tilde{z}_{1:n}, \alpha$), where $\tilde{z}_i = (\tilde{x}_i, \tilde{y}_i)$ (\Cref{eqn:mixup}). 

Finally, we make a practical choice to replace the weights $\{w_i\}_{i=1}^\infty$ sampled from the DP with their expectations, simplifying implementation. 
This choice is further motivated by that randomized weights make little difference in separable settings (see \Cref{sec:equivalency} and \Cref{sec:exp_equiv}). 
The resulting distribution takes the form
\begin{align}
\textstyle \FMMP(\cdot) \propto  \sum_{i=1}^n \Haug_{z_i}(\cdot) + r \Hmixup_{z_{1:n};\alpha}(\cdot), 
\label{eq:mixupDP}
\end{align}
where $r:=c/n$ is the  concentration ratio parameter for some $c\geq0$. 
$\PMMP$ is then the future predictive distribution implied by i.i.d. sampling from $\FMMP$. 

We obtain posterior samples of $\theta$ by repeatedly sampling a sequence of observations from $\FMMP$ and then minimizing the loss $l_\theta$ with respect to this sequence. We note that this sequence is exchangeable by construction, and thus this procedure generates samples from a well-defined martingale posterior. We call this approach \ourmethod{}, and summarize a practical procedure in \Cref{alg:ourmethod}. 

In \Cref{alg:ourmethod}, we split the training sequence into mini-batches,  allowing it to effectively work with a sequence of unbounded length. 
Additionally, we use a fixed dataloader (permuted in the beginning of every training epoch) that cycles through the data rather than randomly sampling. 
While the resulting samples are no longer exact samples from $\FMMP$ (\Cref{eq:mixupDP}), we expect the practical impact  would be minimal.  

\paragraph{Relationship to other methods.}
If the concentration ratio $r=0$ in \Cref{eq:mixupDP}, \ourmethod{} reduces to DE with standard augmentation. If, in addition, $\Haug_{z} := \delta_{z}$, we recover DE without augmentation. If $r=\infty$ in \Cref{eq:mixupDP} and $B=1$ in \Cref{alg:ourmethod}, we recover the original  Mixup  algorithm \citep{zhang2017mixup}. 
And if $r=\infty$ with $B>1$, we recover Mixup Ensemble, as explored by \citet{rahaman2021uncertainty} and \citet{wen2020combining}.

\begin{figure}
    \centering
    \begin{subfigure}[b]{\columnwidth}\includegraphics[width=1\columnwidth]{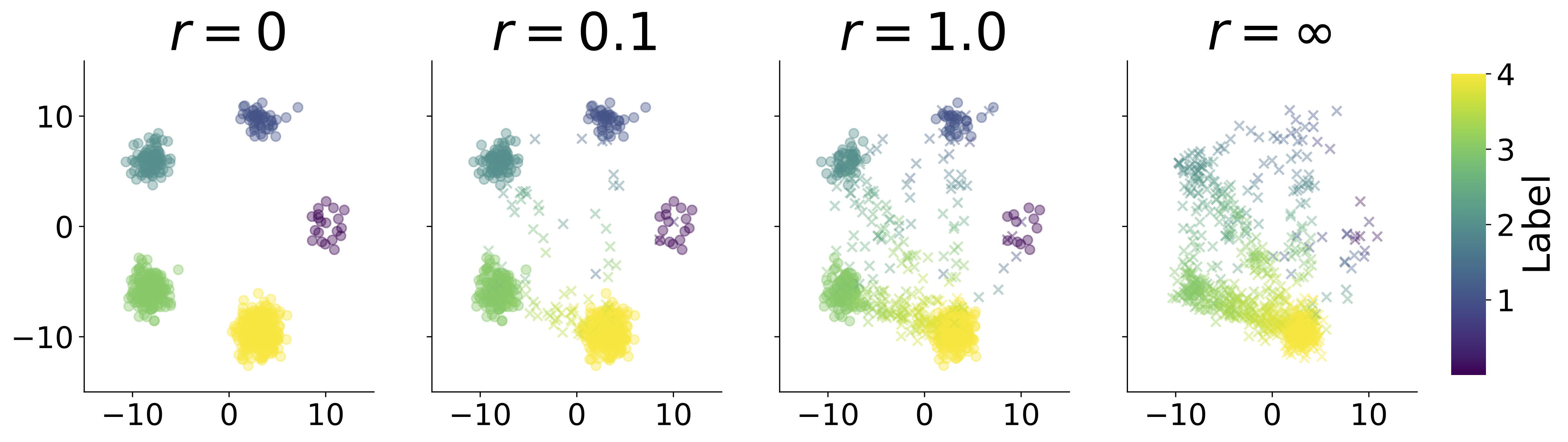}
    \caption{$\FMMP$ sample from \ourmethod{} with varying ratio $r:=c/n$ and $\Haug_x:=\delta_x$ and $\alpha=1.0$. Dots represent samples from observations and crosses are samples from the base measure. The label space is extended to interval $[0,K-1]$.}
    \label{fig:mixupdp-illustraion-data}
    \end{subfigure}
    \begin{subfigure}[b]{\columnwidth}\includegraphics[width=1\columnwidth]{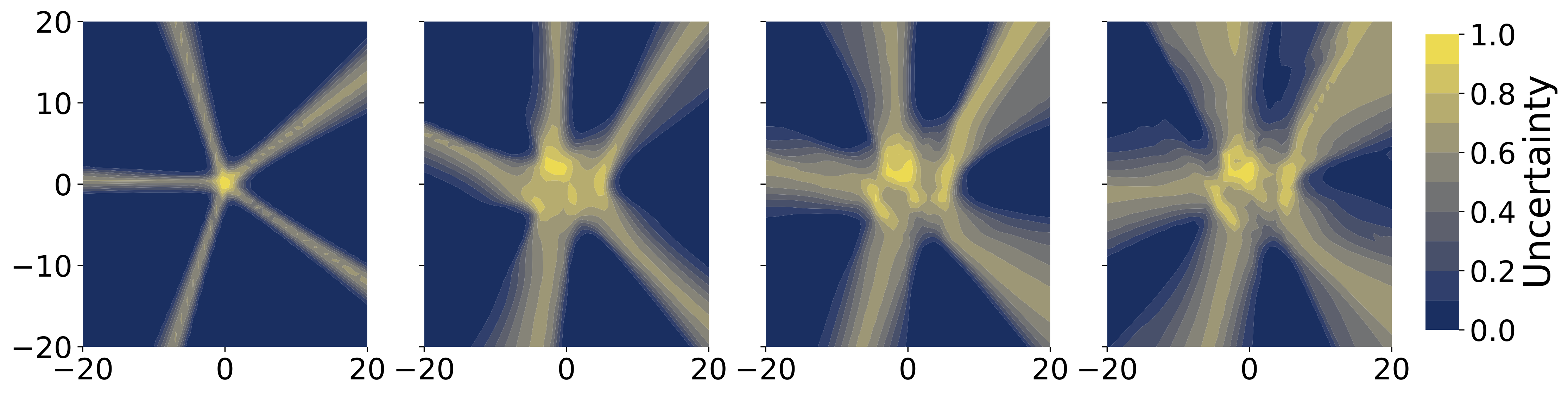}
    \caption{The corresponding predictive uncertainty landscape.}
    \label{fig:mixupdp-illustraion-uncertainty}
    \end{subfigure}
    \caption{Illustration of \ourmethod{} on synthetic classification task $(K=5)$ with $\alpha=1.0$. As  $r$ increases, $\FMMP$ puts more uncertainty on the space between observations, inducing higher predictive uncertainty. }
    \label{fig:mixupdp-illustraion}
\end{figure}

\paragraph{Illustration of the predictive distribution.} We illustrate the effect of the  concentration ratio $r$ in a synthetic 5-class dataset in \Cref{fig:mixupdp-illustraion}. 
When $r=\infty$, all samples from $\FMMP$ are interpolations between observations, as shown in the RHS of \Cref{fig:mixupdp-illustraion-data}. 
However, as we decrease $r$, we see an increasing proportion of samples that are close to the original empirical distribution. 
This behavior avoids over-smoothing  and ensures we are training on sufficient data points that are close to actual observations, while still populating regions between clusters. We can think of  $r$ as describing the extent to which we believe future observations will look like our training data (small $r$) vs the Mixup base measure (large $r$). 

\Cref{fig:mixupdp-illustraion-uncertainty} shows the corresponding uncertainty estimates obtained using \ourmethod{} (see \Cref{app:sec:synthetic} for details). The DE solution ($r=0$) provides low uncertainty across most of the space except around the classification boundary. As we increase $r$, we see increase uncertainty in the regions between the original 5 clusters of observations. 
And Mixup Ensemble ($r=\infty$) provides excessive uncertainty even in the region when observations are present. 

The choice of the Mixup parameter $\alpha$ impacts how close the novel samples from the base measure $H^{\text{Mixup}}$  are to the data. For small values, we tend to see samples close to original observations; for larger values, we see more space-filling behavior. We repeat \Cref{fig:mixupdp-illustraion} in \Cref{app:sec:synthetic} using more values of $\alpha$.

\subsection{Approximate \ourmethod{}
}\label{sec:ourmethod_mc} 

We propose an efficient approximation to \ourmethod{} using MC dropout, where a single neural network is trained with a positive dropout rate using \Cref{alg:ourmethod} (i.e. $B=1$), and dropout is used at test time to generate multiple samples. Since dropout is randomly activated in each training iteration, this procedure implicitly trains an ensemble of models over different draws of data from $\FMMP$ (see \Cref{sec:background}). We refer to this approximation as \ourmethod{}-MC.

%% file: sections/theory.tex
In this section, we show that DE can be viewed as an instance of martingale posteriors, through establishing its equivalency to a finite approximation to the BB in many practical deep learning scenarios; furthermore, we reveal that the predictive distribution underlying such martingale posteriors is mis-specified, motivating the main methodology in the next section.

We first note that both BB and DE are carrying out weighted empirical loss minimization, albeit with different sets of weights (Equations (4) and (5)).
Recent theoretical work has shown that, if training data is separable and under certain conditions on the neural network, loss, and optimizer, a neural network will converge to the L2 max margin solution \citep{lyu2019gradient,wei2019regularization,nacson2019convergence,chizat2020implicit}. Further, \citet{xu2021understanding} have shown that this margin is invariant to replacing the empirical loss $\sum_i \ell_\theta(z_i)$ with a weighted loss $\sum_i w_i\ell_\theta(z_i)$, where the $w_i$ are bounded weight values. 


Combining these observations leads to the following proposition:


\begin{proposition}\label{prop:equiv}
    (Informal) If a dataset $z_{1:n}$ is separable under a given homogeneous neural network with parameters $\theta$, trained via stochastic gradient descent using an exponentially tailed loss (e.g.\ cross-entropy) with weak regularization, then
    any posterior sample of $\theta$ obtained via an appropriately stabilized version of BB could also have been obtained via DE, and vice versa. 
\end{proposition}
See the formal statement, assumptions and proof, plus definition of stabilized BB, in \Cref{appendix-sec:proof}.

This result extends previous empirical work \citep{nixon2020bootstrapped} that shows the frequentist bootstrap \citep[FB,][]{efron1992bootstrap} underperforms DE in deep learning applications. The authors attributed FB's underperformance to the fact that each bootstrap sample only contains 63.7\% of the unique observations from the training data, undermining the generalization power of the learned model. 

In BB, since all observations are represented in every bootstrap sample, one might expect to avoid the pitfall of FB, and even outperform DE. However, \Cref{prop:equiv} suggests that it is not the case, as BB is functionally equivalent to DE. In fact, \Cref{sec:exp_equiv} shows that there is little difference in their empirical performance.

\paragraph{Mis-specification of BB and DE.}
The success of BB 
relies on the assumption that the predictive distribution of future observations can be well approximated by the empirical distribution. Clearly this assumption is not true: the test set will not include only repeats of the training data. In an underparametrized model,  
this mis-specification can typically be ignored; since a perfect training accuracy is rarely achieved, differently weighted  training set causes the model to focus on different regions, effectively capturing some uncertainty in the input space.

Conversely, BB's assumptions are fundamentally mis-specified in the deep learning context. 
Overparameterized deep neural networks perfectly fits all training observations, and hence different weights do not impact their convergent solutions (in the asymptotic regime). However, realistic test cases involve novel observations, and including these observations to the training set which \textit{will} impact the convergent solutions of models. For this reason, we argue that BB --- and by extension of \Cref{prop:equiv}, DE --- correspond to mis-specified versions of martingale posteriors.

%% file: sections/relatedworks.tex
\section{Related works}
Several works have applied bootstrapping-type methods to deep learning, some of which fall under the definition of martingale posteriors. \citet{osband2016deep} use a frequentist bootstrap to perform inference in deep Q-networks (DQN), improving performance over a single DQN model. \citet{shin2021neural} use a hypernetwork approach to approximately implement BB in neural networks. \citet{newton2021weighted} and \citet{osband2018randomized} modify the BB to incorporate a randomized prior term in the loss function. \citet{lee2023martingale} specifically follow a MP approach, specifying $\mathbb{P}_\infty$ using an exchangeable generative neural network to quantify uncertainty for neural processes \citep{garnelo2018neural}; however their approach is specific to neural processes \citep{garnelo2018neural} rather than general neural networks. 
By comparison, \ourmethod{} is data-driven and applies to general architectures. 

More broadly, several approximate inference methods for BNNs have been proposed. In addition to DE and MC Dropout (discussed in \Cref{sec:background}), options include Monte Carlo methods \citep{mackay1992practical,chen2014stochastic,zhang2019cyclical}; variational inference \citep{graves2011practical,blundell2015weight}; Laplace approximations \citep{daxberger2021laplace}; and expectation propogation \citep{hernandez2015probabilistic}. With the exception of DE and Monte Carlo methods, these approaches only explore a single mode of the posterior. Sampling from multiple modes has been found to improve the quality of posterior estimates \citep{wilson2020bayesian}, and the posterior estimate obtained using DE has been shown to be closer to a ``gold standard'' Monte Carlo estimate than variational approaches \citep{izmailov2021bayesian}. 

When $r=\infty$, \ourmethod{} corresponds to a ensemble of models trained using Mixup (which we refer to as Mixup Ensemble). This setting has been shown to yield underconfident predictions \citep{wen2020combining,rahaman2021uncertainty}. Calibration-adjusted Mixup \citep[CAMixup,][]{wen2020combining} 
is a modification of Mixup Ensemble that aims to reduce this underconfidence  by only applying mixup to classes where the model is already over-confident (as assessed on a validation set after each training epoch). Consequently, all members of CAMixup need to be trained simultaneously. In contrast, our approach allows individual training of each model, alleviating memory constraints. 









%% file: sections/experiments.tex
\section{Experiments}\label{sec:experiments}
In this section, we provide empirical results supporting the equivalency of BB and DE, and empirically evaluate the performance of \ourmethod{}. For code, see \url{https://github.com/apple/ml-MixupMP}. Experiments were carried out using Apple internal clusters.

When comparing BB and DE, we look at two datasets: MNIST \citep{lecun1998gradient} and FMNIST \citep{xiao2017fashion}, and do not use any data augmentations. For the analysis of \ourmethod{}, we look at three datasets: CIFAR10, CIFAR100 \citep{krizhevsky2009learning}, and FMNIST. For the CIFAR datasets, we specify $\Haug$ using the augmentations {\tt RandomResizeCrop} and {\tt RandomHorizontalFlip} in PyTorch \citep{Paszke2019PyTorchAI}; we also use these augmentations in competing methods. For FMNIST, we do not use augmentations since the images are centered and standardized. 
Unless otherwise stated, we use $B=4$ ensemble members for all DE, Mixup Ensemble, BB and \ourmethod{} experiments to estimate the posterior mean, and $B=20$ for implicit ensemble methods based on MC Dropout, 
following \citet{wen2020combining}. 
We include additional implementation details in \Cref{app:equivalency} and \Cref{app:extra_ourmethod_results}. 


\paragraph{Evaluation Metrics.} 
We use 0-1 accuracy (ACC) and negative log likelihood (NLL) to evaluate predictive performance. For uncertainty quantification, we consider  expected calibration error \citep[ECE,][]{naeini2015obtaining}, over-confidence error (OE), and under-confidence error (UE); see \Cref{app:uq_metrics} for details. Finally, we assess the predictive entropy as a predictor for  uncertainty in the event of distribution shift. 

\subsection{Equivalency of BB and DE}\label{sec:exp_equiv}

In Section \ref{sec:equivalency}, we argued that, on separable datasets, BB is equivalent to DE. Here, we demonstrate this claim empirically. We compare test set performance on MNIST and FMNIST under BB and DE, each with $B=4$ simulations. In every simulation, the model is run until it achieves 100\% training accuracy, plus 1000 further epochs, to ensure a separable setting is achieved. The initializations were kept fixed between the two approaches,  either from a random initialization, or from a separately pretrained DE. 

From \Cref{tab:BBcomp}, we see little difference between the two methods in terms of test accuracy and ECE when the initialization is set to a good solution provided by a separately pretrained DE. Moreover, in \Cref{app:equivalency}, we show that the individual ensemble members within BB and DE do not significantly differ in this scenario.

When DE and BB are randomly initialized, there is a small difference between the two. We note that \Cref{prop:equiv} only concerns the convergent behavior; it is possible that the weights added in BB nudge the models to different basins of attraction during early-stage training dynamics. We may be also seeing differences in convergence rate. This result supports \citet{byrd2019effect}, who find that, while the effect of importance weighting will ultimately vanish as training progresses, empirically ``models with more extreme weighting converge more slowly in decision boundary''. In our case, it could take BB  longer to converge due to variation among the random weights, as composed to the uniform weights used by DE.



\begin{table}
\caption{Comparing BB with DE. Models are either randomly initialized (same set of seeds for DE and BB), or initialized from a pretrained DE. 
}
\resizebox{\columnwidth}{!}{%

\input{tables/bb_vs_de/bb_vs_de}

}
\label{tab:BBcomp}
\end{table}

\begin{figure*}[!t]
    \centering
    \includegraphics[width=\textwidth]{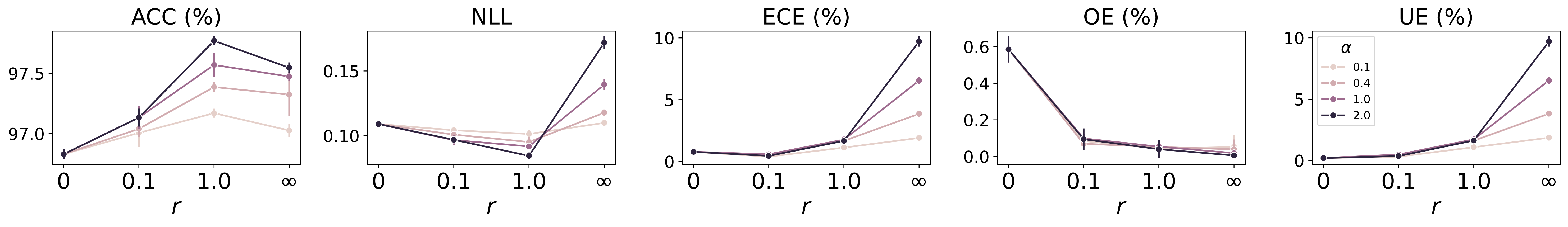}
    \caption{Impact of $\alpha$ and $r$ on test set performance of \ourmethod{} on CIFAR10. $r=0$ corresponds to DE; $r=\infty$ corresponds to Mixup Ensemble. Results for CIFAR100 and FMNIST are included in \Cref{app:extra_ourmethod_results}.} 
    \label{fig:ablation-cifar10}
\end{figure*}

\subsection{Ablation study: Impact of hyperparameters in \ourmethod{}}\label{sec:ablation}
The performance of \ourmethod{} hinges on the ability of $\FMMP$ to capture uncertainty about future data. 
This ability 
depends on the choices of random augmentations in $\Haug$, the Mixup parameter $\alpha$ used in Mixup, and the concentration ratio $r$  
of Mixup samples to (augmented) observations. 

In \Cref{fig:ablation-cifar10}, we look at the performance of \ourmethod{} on CIFAR10, across various values of $\alpha$ and $r$. 
Note that when $r=0$ we recover DE (which does not rely on $\alpha$), and when $r=\infty$, the Mixup Ensemble. 

For any fixed $\alpha$, we observe \ourmethod{}'s predictive performance (ACC and NLL) with moderate value of $r$ improves upon both the $r=0$ and $r=\infty$ solutions (i.e., DE and Mixup Ensemble).  For uncertainty calibration, \ourmethod{} with  $r=0.1$ achieves slightly better ECE compared to DE ($r=0$), but larger $r$ can greatly inflate the ECE, especially for Mixup Ensemble ($r=\infty$). 
This poor calibration performance of Mixup Ensemble also corroborates previous findings in \citet{wen2020combining}. Looking at OE and UE, we observe that increasing $r$ tends to reduce the model's over-confidence but boosting the under-confidence. This result is expected since we are increasing the relative importance of uncertain training examples.


On the other hand,  for moderate $0<r<\infty$, raising $\alpha$ generally leads to improvements in ACC and NLL, with only a marginal increase in ECE. However for Mixup Ensemble ($r=\infty$), large values of $\alpha$ can significantly hurt NLL and ECE, despite enhancing the accuracy. In theory, increasing $\alpha$ will shift the Mixup base measure to higher uncertainty region, with $\alpha \to \infty$ leading to maximum  uncertainty since pairs of observations are equally mixed. Such effect of large $\alpha$ can be  amplified with large $r$, explaining the uncertainty calibration behavior of Mixup Ensemble. 

We conclude that the optimal value of $r$ should be somewhere between the extremes of $0$ and $\infty$. DE ($r=0$) assumes that \textit{all} future data are copies of previous observations, which leads to overconfident estimates and worse accuracy on new data points. Conversely, Mixup Ensemble ($r=\infty$) assumes that \textit{all} future data are uncertain, leading to underconfidence despite an improved accuracy. \ourmethod{} interpolates between these two extremes, balancing the predictive performance and uncertainty calibration. We include additional analysis on CIFAR100 and FMNIST in \Cref{app:extra_ourmethod_results} where we observe similar trends. 

\begin{table*}[!t]
\caption{Performance comparison on three datasets. 
Bolded metrics are the best (within 2 standard errors, omitted for space) in each group of \{single model, explicit ensemble, implicit ensemble\}; $*$ indicates the best  among all methods. Single model results are averaged over 6 independent runs and  ensemble results are averaged over 3 independent ensembles. 
We include all ablation results with standard errors in \Cref{app:extra_ourmethod_results}}\label{tab:results-indist}
\resizebox{\textwidth}{!}{%
\input{tables/comparison/test_ind_alpha2.0_semFalse}

}
\end{table*}

\subsection{Comparison with other uncertainty quantification methods}
\label{sec:comparison}

Next, we evaluate how \ourmethod{} performs relative to standard neural networks, and to other Bayesian and ensemble-based methods.  
These methods include (1) \textit{ NN}, a single neural network; (2) \textit{Mixup},  a single neural network trained with  Mixup augmentation; (3) \textit{Laplace}, a Laplace approximation to the BNN posterior \citep{daxberger2021laplace}; (2) \textit{MC Dropout}; (3) \textit{DE}; 
(4) \textit{Mixup Ensemble};  
(5)  \textit{CAMixup-MC}, an efficient version of CAMixup \citep{wen2020combining}.\footnote{While \citet{wen2020combining} propose two other ensembling versions of CAMixup,  they either impose significant  memory constrainst, or require modifications to the underlying neural network.} 
See \Cref{app:extra_ourmethod_results} for details. Additionally, we consider \textit{\ourmethod{}-MC}, an efficient approximation to \ourmethod{} as introduced in \Cref{sec:ourmethod_mc}. Similarly we include  \textit{Mixup-MC} as an approximation to Mixup Ensemble (equivalent to \ourmethod{}-MC with $r=\infty$). For all the methods that use Mixup augmentation, we set $\alpha=2.0$. 
We refer to  the class of models that form a distribution over parameters as \textit{probabilistic} models. All above methods methods except for single NN and Mixup are probabilistic. 
The results are summarized in \Cref{tab:results-indist}.


We first note that single NN underperforms probablistic models in almost all cases, 
supporting findings from earlier works \citep{izmailov2021bayesian,ovadia2019can}. 
Among the probabilistic models, we find that \ourmethod{} with $r=1.0$ outperforms  Laplace,  MC Dropout, DE and Mixup Ensemble in terms of accuracy and NLL, either using explicit ensemble or implicit ensemble. \ourmethod{} with $r=0.1$ performs slightly worse than $r=1.0$ on these two  metrics, but is better than other approaches in ECE. In terms of uncertainty calibration, \ourmethod{} with $r=0.1$ is the best or close to the best among all methods. 

We next examine the performance of \ourmethod{}'s more efficient variant, \ourmethod{}-MC. While \ourmethod{}-MC performs slightly worse than \ourmethod{}, it outperforms  MC Dropout, Laplace, and Mixup-MC on all metrics. 
CAMixup-MC achieves good calibration performance---its primary goal---as it has a separate validation set to adjust calibration. However, CAMixup-MC underperforms \ourmethod{} and \ourmethod{}-MC in terms of accuracy and NLL. These observations highlight that \ourmethod{}-MC can serve as an efficient proxy with little performance compromise. 

\subsection{Robustness to distribution shift}
\label{subsec:robustness}

Lastly, we evaluate the generalization power and the reliability of uncertainty estimates of \ourmethod{} in distribution shift settings. 
We consider the CIFAR10-C dataset which contain 19 types of corruptions to the original CIFAR10 test set, each with 5 shift intensity levels \citep{hendrycks2018benchmarking}. 

\begin{figure}[h]
    \centering
    \begin{subfigure}[b]{\columnwidth}
\includegraphics[width=\linewidth]{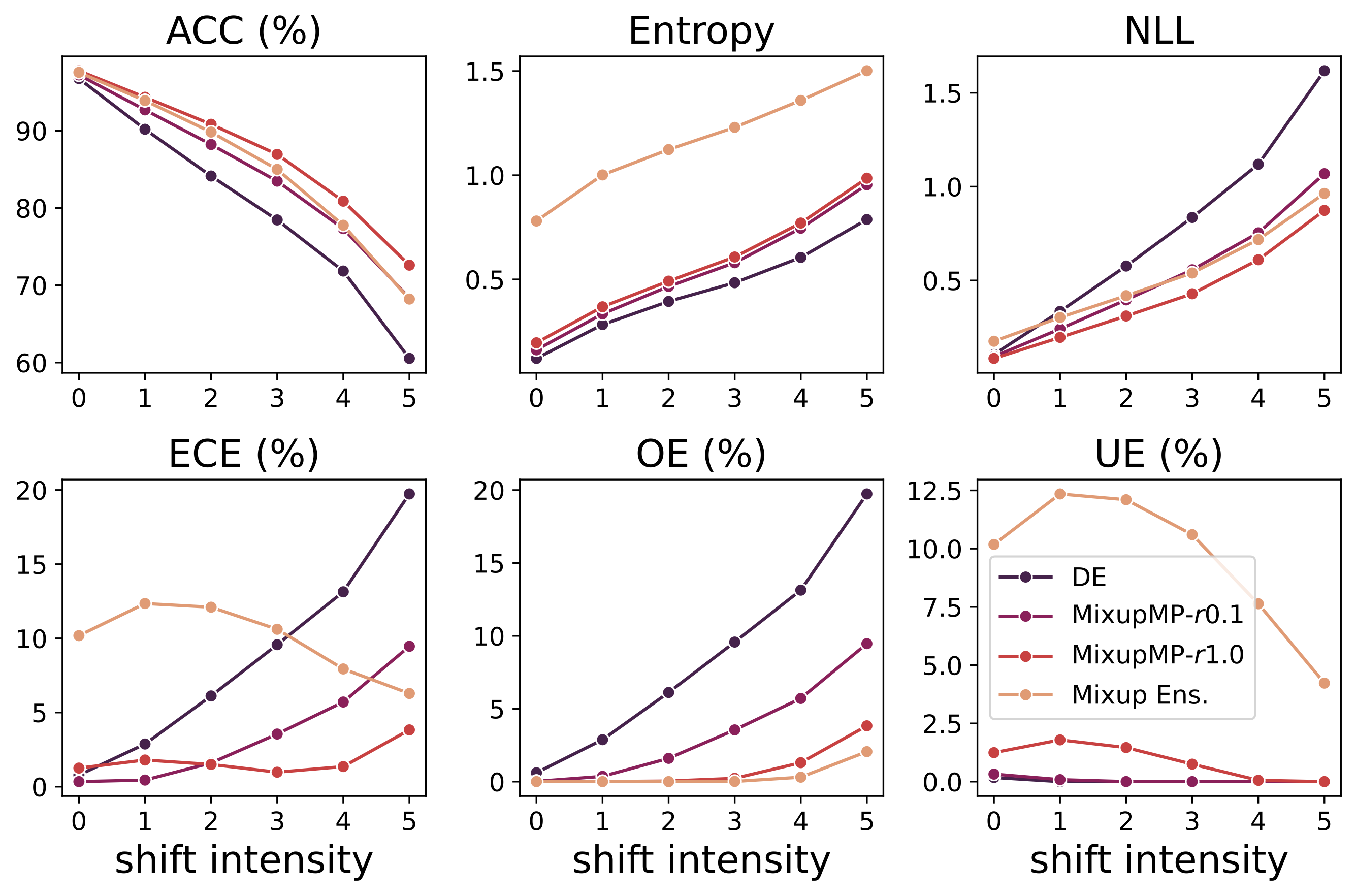}
\end{subfigure}
\caption{Performance under distribution shift using CIFAR10-C dataset. The distribution shift intensity ranges from 0 to 5, where 0 indicates no shift.}
\label{fig:cifar10-robustness}
\end{figure}
In \Cref{fig:cifar10-robustness}, we compare the performance of DE, \ourmethod{}, and Mixup Ensemble, each with the Mixup parameter $\alpha=2$. 
 For all methods, as the test data shift intensity increases, both ACC and NLL deteriorate as expected; however, performance deteriorates less for \ourmethod{} and Mixup Ensemble. There is a concurrent increase in the predictive entropy for all models,  suggesting that their uncertainty estimates are informative. In particular, the higher the concentration ratio $r$ for \ourmethod{}, the higher the predictive entropy for the same shift intensity (recalling that DE corresponds to \ourmethod{} with $r=0$ and Mixup Ensemble corresponds to $r=\infty$). 
 
Regarding calibration, in alignment with our findings in the in-distribution setting in \Cref{sec:ablation}, DE tends to be over-confident and Mixup Ensemble tends to be under-confident. Meanwhile, \ourmethod{} with moderate values of $r$ avoids both pitfalls, achieving better calibration in almost all cases.

Finally, we highlight that \ourmethod{} with $r=1.0$ achieves the best accuracy, NLL and ECE across all shift intensity levels in \Cref{fig:ablation-cifar10}. We include additional results on CIFAR100-C dataset \citep{hendrycks2018benchmarking} and comparison to other methods in \Cref{app:extra_ourmethod_results}. Through this comprehensive study, we show that \ourmethod{} achieves superior performance in various distribution shift settings.

%% file: tables/bb_vs_de/bb_vs_de.tex
\begin{tabular}{ccccc}
    \hline 
    Dataset & Method & Init. & Acc (\%) & ECE (\%) \\
    \midrule
\multirow{4}{*}{MNIST} & DE & random & 99.33 & 0.41 \\
 & BB & random & 99.17 & 0.24 \\ 
 & DE & DE & 99.33 & 0.40  \\ 
 & BB & DE & 99.33 & 0.41 \\ 
 \midrule
\multirow{4}{*}{FMNIST} & DE & random & 91.52 & 2.32\\ 
 & BB & random & 91.21 & 2.01 \\ 
 & DE & DE & 91.57 & 2.42 \\ 
 & BB & DE & 91.55 & 2.37\\ 
 \hline
\end{tabular}

%% file: tables/comparison/test_ind_alpha2.0_semFalse.tex
\begin{tabular}{llll|lll|lll} \hline 
\multicolumn{1}{l|}{Dataset}& \multicolumn{3}{l|}{CIFAR10}& \multicolumn{3}{l|}{CIFAR100}& \multicolumn{3}{l}{FMNIST} \\ \hline 
\multicolumn{1}{l|}{Metric} & ACC (\%)  & NLL  & ECE (\%)  & ACC (\%)  & NLL  & ECE (\%)  & ACC (\%)  & NLL  & ECE (\%) \\ \hline 
\multicolumn{1}{l|}{\textbf{Single model}}  &  &  &  &  &  &  &  &  & \\ 
\multicolumn{1}{l|}{NN} &$96.12$ &$0.1473$ &$2.02$ &$80.82$ &$0.7816$ &$\mathbf{4.78}$ &$93.72$ &$0.2181$ &$2.89$\\ 
\multicolumn{1}{l|}{Mixup ($\alpha$=2.0)} &$\mathbf{96.88}$ &$0.1854$ &$8.59$ &$\mathbf{81.83}$ &$\mathbf{0.7614}$ &$8.54$ &$\mathbf{94.15}$ &$\mathbf{0.1948}$ &$\mathbf{1.46}$\\ 
\multicolumn{1}{l|}{Laplace} &$96.04$ &$\mathbf{0.1344}$ &$\mathbf{0.80}$ &$80.95$ &$1.0107$ &$22.96$ &$89.67$ &$0.4088$ &$5.89$\\ 
\hline 
\multicolumn{1}{l|}{\textbf{Explicit ensemble (B=4)}}  &  &  &  &  &  &  &  &  & \\ 
\multicolumn{1}{l|}{DE} &$96.83$ &$0.1090$ &$0.78$ &$83.28$ &$0.6413$ &$\mathbf{3.14}$ &$94.30$ &$0.1768$ &$1.36$\\ 
\multicolumn{1}{l|}{MixupMP (r=0.1, $\alpha$=2.0)} &$97.13$ &$0.0969$ &$\mathbf{0.46}^*$ &$84.46$ &$0.6127$ &$3.95$ &$\mathbf{94.75}^*$ &$\mathbf{0.1610}^*$ &$1.27$\\ 
\multicolumn{1}{l|}{MixupMP (r=1.0, $\alpha$=2.0)} &$\mathbf{97.77}^*$ &$\mathbf{0.0845}^*$ &$1.67$ &$\mathbf{85.98}^*$ &$\mathbf{0.5548}^*$ &$5.49$ &$\mathbf{94.70}^*$ &$0.1662$ &$\mathbf{1.01}$\\ 
\multicolumn{1}{l|}{Mixup Ensemble ($\alpha$=2.0)} &$97.55$ &$0.1717$ &$9.71$ &$84.48$ &$0.6768$ &$13.30$ &$\mathbf{94.74}^*$ &$0.1776$ &$2.47$\\ 
\hline 
\multicolumn{1}{l|}{\textbf{Implicit ensemble (B=20)}}  &  &  &  &  &  &  &  &  & \\ 
\multicolumn{1}{l|}{MC Dropout} &$96.16$ &$0.1315$ &$1.46$ &$80.83$ &$0.7451$ &$3.69$ &$94.41$ &$0.1806$ &$1.90$\\ 
\multicolumn{1}{l|}{MixupMP-MC (r=0.1, $\alpha$=2.0)} &$96.58$ &$0.1145$ &$\mathbf{0.42}^*$ &$82.42$ &$0.7079$ &$3.39$ &$\mathbf{94.72}^*$ &$0.1651$ &$1.46$\\ 
\multicolumn{1}{l|}{MixupMP-MC (r=1.0, $\alpha$=2.0)} &$\mathbf{97.27}$ &$\mathbf{0.1005}$ &$1.21$ &$\mathbf{83.58}$ &$\mathbf{0.6419}$ &$4.21$ &$\mathbf{94.81}^*$ &$\mathbf{0.1619}^*$ &$\mathbf{0.97}^*$\\ 
\multicolumn{1}{l|}{Mixup-MC ($\alpha$=2.0)} &$96.80$ &$0.2084$ &$10.98$ &$81.75$ &$0.7616$ &$10.71$ &$94.69$ &$0.1796$ &$2.84$\\ 
\multicolumn{1}{l|}{CAMixup-MC ($\alpha$=2.0)} &$96.11$ &$0.1365$ &$1.21$ &$80.19$ &$0.7780$ &$\mathbf{2.14}^*$ &$94.27$ &$0.1818$ &$\mathbf{1.04}^*$\\ 
\hline 
 \end{tabular}

%% file: sections/discussion.tex
\section{Discussion}
In this work, we show that the posterior distribution implied by deep ensembles can be framed as a martingale posterior, but caution that when viewed in this light, it is misspecified in most settings. Instead, we propose \ourmethod{}, a novel martingale posterior approach that uses state-of-the-art data augmentation techniques to better captures predictive uncertainty in image data, leading to improved predictive performance and uncertainty quantification. We hope that this work will spark increased interests in predictive approaches to quantify uncertainty for deep learning, and inspire future work on other structured data modalities.

%% file: sections/appendix.tex
\onecolumn
\aistatstitle{Supplementary materials: A Predictive View of Uncertainty Quantification in Deep Learning}

\appendix 

\renewcommand\thesection{\Alph{section}} 

\counterwithin{table}{section}
\counterwithin{figure}{section}

\section{Formal statement and proof of \Cref{prop:equiv}}\label{appendix-sec:proof}

\subsection{Both DE and stabilized BB converge to the max margin solution}
Several works have considered the limiting margin behavior of homogeneous neural networks \citep{wei2019regularization,lyu2019gradient,ji2020directional, xu2021understanding}. The margin for a single datapoint $z_i=(x_i, y_i)$ is defined as $\gamma_i = y_i f_\theta(x_i)$,  and the margin for the entire dataset as $\gamma_{\min} (\theta) = \min_i \gamma_i$. In the case of $L$-homogeneous neural networks---i.e., networks where \ $f_{c\theta}(x) = c^L f_\theta(x)$ for some $L>0$ and all $c>0$---the margin $\gamma_{\min}(\theta)$ scales linearly with $||\theta||_2^L$, so we consider the normalized margin,

$$\tilde{\gamma}(\theta) = \gamma_{\min}\left(\frac{\theta}{||\theta||_2}\right) = \frac{\gamma_{\min}(\theta)}{||\theta||_2^L}$$

Under certain conditions, this normalized margin has been shown to converge to a max-margin solution $\gamma^*$ \citep{wei2019regularization,lyu2019gradient,ji2020directional}. \citet{xu2021understanding} show that such behavior can also be found when we incorporate weights in empirical loss minimization, where we have a loss of the form form $\mathcal{L}(\theta) = \sum_i w_i \ell_\theta(z_i) + \lambda||\theta||^r$ for some $r>0$ and $\lambda\rightarrow 0$, referred to as weak regularization. 

Below, we consider the binary classification setting, with $y_i \in \{-1, +1\}$. 
Extension to the multi-class setting is straightforward. 
Following \citet{xu2021understanding}, we make the following assumptions:

\paragraph{Assumptions}
\begin{enumerate}[label=A\arabic*]
\item The loss takes the form $\ell_\theta(z_i) = \ell(y_i  f_\theta(x_i))$, where $\ell(\cdot)$ has exponential tail behavior s.t.\ $\lim_{u\rightarrow \infty}$$ \ell(-u) = \lim_{u\rightarrow \infty} \Delta \ell(-u) = 0$.
\item $f_\theta(x_i)$ is $L$-homogeneous, i.e.\ $f_{c\theta}(x) = c^L f_\theta(x)$ for some $L>0$, all $c>0$, and all $x$ and $\theta$.
\item $f_\cdot(x)$ is $\beta$-smooth and $l$-Lipschitz on $\mathbb{R}^d$ for all $x$.
\item The data are separable by $f_\theta$, and this condition can be reached via gradient descent. Further, $y_i f_{\theta^*}(x_i) \geq \gamma^* > 0$ for each $i$.
\item The weights $w_i$ are bounded, s.t.\ $w_i \in [1/M, M]$ for some positive $M$.
\end{enumerate}

\begin{lemma}[Proposition 3 of \citet{xu2021understanding}]
If Assumptions A1-A5 hold, then $\lim_{\lambda \rightarrow 0} \tilde{\gamma}(\theta) \rightarrow \gamma^*$, i.e.\ $f_\theta$ converges to the max margin solution.\label{lem:xu}
\end{lemma}

Cross-entropy loss meets assumption A1; see \citet{xu2021understanding} for more general sufficient descriptions of exponential tail behavior. Assumption A2 is met by feedforward and convolutional neural networks with ReLU activations and no bias terms. Assumption A4 is common in image datasets where we can achieve perfect train set classification. This is often the case in practice; for example, the MNIST and FashionMNIST training datasets used in this paper can be perfectly classified using a CNN with no bias terms.

DE trivially meets Assumption A5 since the weights are all $1/n$. In the Bayesian bootstrap setting, we ensure assumption A5 is met via the following definition of stabilized BB weights:

$$\begin{aligned}
    (\tilde{w}_1,\dots, \tilde{w}_n) \sim& \mbox{Dirichlet}(1,\dots, 1)\\
    w_i =& \frac{\tilde{w}+\eta}{n+\eta},
\end{aligned}$$
where $\eta = 1/(M-n)$ for some $M>n$.

It therefore follows from Lemma 1 that both DE and stabilized BB converge to the max margin classifier, provided assumptions A2, A3 and A4 are met. As noted by \citet{xu2021understanding}, in practice the smoothness condition in A3 is not met when using ReLU activations; however, they find that in practice using ReLU rather than a smoother activation function makes little difference. Similarly, incorporating bias terms in a CNN with ReLU activations violates the homogeneity assumption A2; however in practice we found little difference when bias terms are included.

\subsection{Equivalency of DE and stabilized BB}

We restate \Cref{prop:equiv} more formally:

\setcounter{proposition}{0}
\begin{proposition} 
    Assume A1-5 are met, and inference is run until convergence of the normalized max margin. Then, any posterior sample obtained via stabilized BB could also have been obtained via DE, and vice versa.
\end{proposition}
This is a direct consequence of \Cref{lem:xu}: Any classifier that minimizes the randomly weighted loss used in stabilized BB will also minimize the unweighted loss.

\paragraph{Remark} \Cref{prop:equiv}
 implies that a sample from DE is a valid minimizer of stabilized BB. This is validated empirically in \Cref{sec:exp_equiv} and \Cref{app:equivalency}: when we initialize a BB sample using a pretrained DE sample, we do not see any significant change in the resulting classifier. 
 
 However, \Cref{prop:equiv} does not consider the probability of obtaining such a sample under the two schemes, when trained from a random initialization. \Cref{lem:xu} only considers asymptotic behavior of the margin on the training set; different weighting schemes will likely impact early-stage learning behavior which can impact \textit{which} local minima the optimization ends up in. Indeed, \citet{xu2021understanding} show that weights can have an impact on out-of-sample performance in the finite-sample setting. In principle, this impact could lead to differing distributions of fitted parameters under DE and BB.

 However, our experiments in \Cref{sec:exp_equiv} and \Cref{app:equivalency} indicate that there is not a significant practical difference. When BB samples are initialized using the same random initialization as DE samples, we do see some variation in the resulting classifiers as a result of the different training dynamics; however the differences are small. This finding is in line with previous empirical work that shows importance weighting has no impact on the limiting behavior of neural networks, under exponentially-tailed losses \citep{byrd2019effect,wang2021importance}.


We note that this result does not necessarily imply that the distribution over solutions will be the same under both paradigms: the weights before the loss terms will likely impact early-stage learning behavior which can impact \textit{which} local minima the optimization ends up in. Indeed, \citet{xu2021understanding} show that weights can have an impact on out-of-sample performance in the finite-sample setting.

However, empirically we find there to be very little difference in the resulting behavior of neural networks trained under Bayesian bootstrapping and deep ensembles (\Cref{tab:BBcomp}/\Cref{tab:BBcomp_extended} and \Cref{fig:seed_comp}). This finding is also supported by empirical work that shows importance weighting has no impact on the limiting behavior of neural networks, under exponentially-tailed losses \citep{byrd2019effect,wang2021importance}.

\begin{figure}[!t]
\centering
    \begin{subfigure}[b]{0.55\columnwidth}\includegraphics[width=\linewidth]{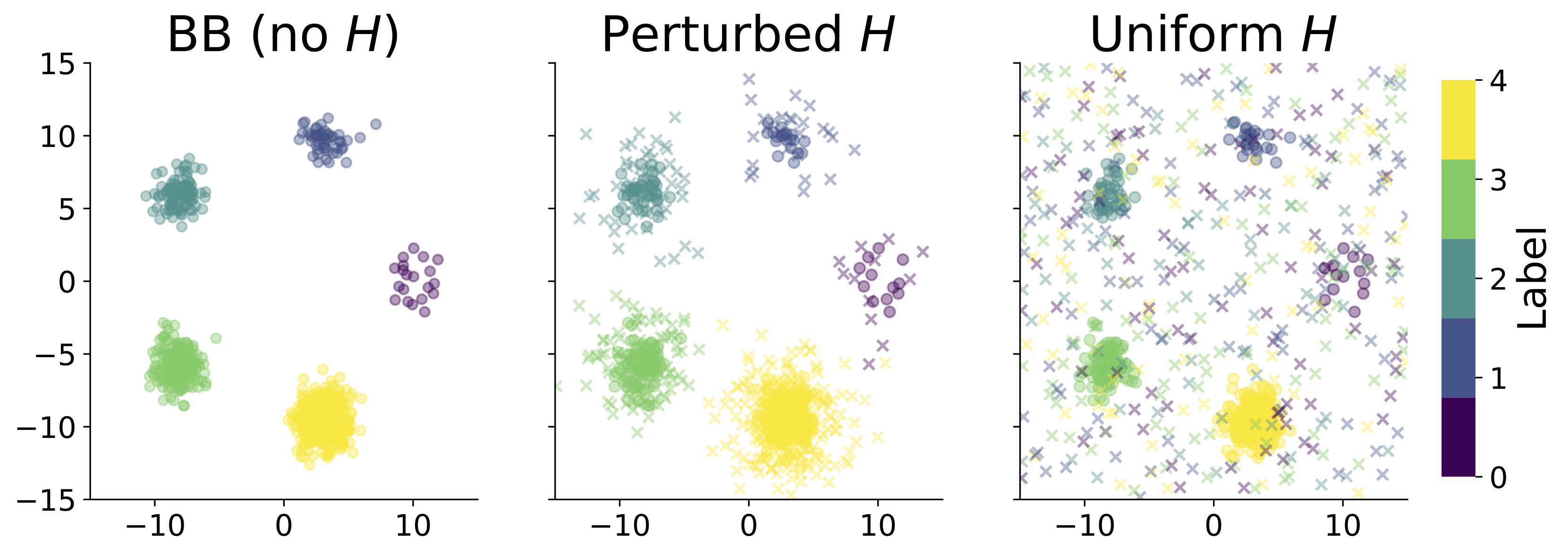}
    \caption{$\FMMP$ sample from different MGs. Left: Bayesian bootstrap (BB); Middle: DP-MP with the base measure $H$ being a perturbed distribution; Right: DP-MP with $H$ being a uniform measure. Dots represent samples from observations and crosses are samples from the base measure. The label indexing starts from $0$ to $K-1$.}
    \label{fig:mp-illustraion-data}
    \end{subfigure}
    \begin{subfigure}[b]{0.58\columnwidth}\includegraphics[width=\linewidth]{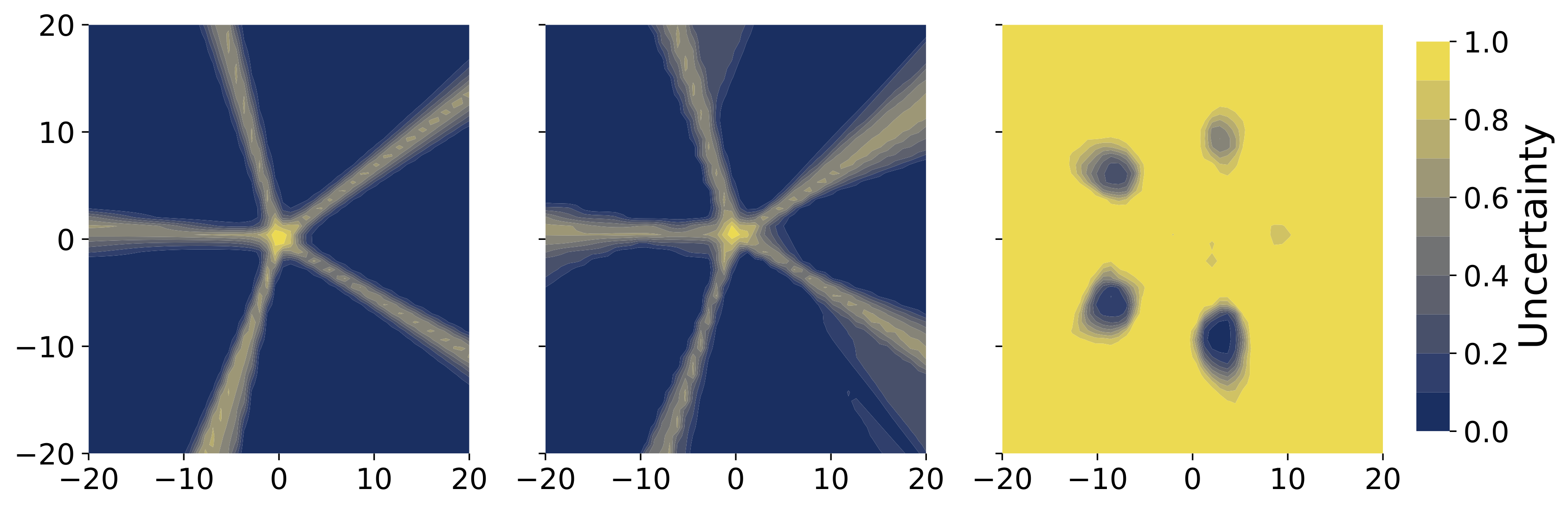}
    \caption{The corresponding predictive uncertainty landscape. BB and DP-MP with perturbed $H$ both have uncertainty centered around decision boundary, with the later has greater uncertainty; DP-MP has high uncertainty in region where observations are absent or scarce.}
    \label{fig:mp-illustraion-uncertainty}
    \end{subfigure}

    \caption{Illustration of Martingale posteriors. Different specifications of the future predictive distribution lead to different uncertainty quantification behaviors.}
    
    \label{fig:mp-illustraion}
\end{figure}

\section{Illustrations of Martingale posteriors and synthetic experiment details}
\label{app:sec:synthetic}

\subsection{Illustration of martingale posteriors}
\label{app:subsec:illustration-of-mp}

We use a 2D synthetic experiment to illustrate how different specifications of future predictive distributions $\mathbb{P}_\infty$ can lead to different uncertainty quantification behaviors.

We randomly generate 5 clusters of 2D inputs of size $[20, 50, 100, 200, 500]$, each corresponding to a class $k\in[1,2,3,4,5]$. We consider three specifications of $\mathbb{P}_\infty$, all through specifying an exchangeable latent distribution $F_\infty$ (see \Cref{sec:bg_mp}): (1) Bayesian Bootstrap (BB), (2) DP-MP with $c=1$ and base measure $H=$ ``Perturbed" -- a data-driven model that injects random noise to observed inputs and preserves their labels, and (3) DP-MP with $c=1$ and $H=$ ``Uniform" -- an uninformative prior model that uniformly samples inputs and labels.

More specifically, in (2) the perturbed $H$ generates a new pseudo sample $z'=(x',y')$ by choosing a random observation $z=(x,y)$ and then set $x':=x+\epsilon$ with $\epsilon \sim \mathcal{N}(0, 4)$ and $y':=y$; and in (3) the uniform $H$ generates independent uniform pseudo samples $z'=(x',y')$ where $x'\sim \mbox{Uniform}[-15,15]^2$, and $y' \sim \mbox{Categorical}(1,2,3,4,5)$.

We use a feed forward neural network with the hidden units size of $[16,32,16]$. 
We obtain their posterior samples $\{\theta^{(b)}\}_{b=1}^B$ according to \Cref{alg:dp-mp} with $B=10$ and $T=n$, the total size of observations. In each simulation, the network is trained from a random initialization with a learning rate of 0.5 for 10,000 epochs, and no mini-batch is used.

The predictive uncertainty is computed as $1-\sum_{k=1}^{K} p^2 (y=k|z)$ (as suggested by \citet{abe2022deep}) and is re-scaled to $[0,1]$, where $p(\cdot|z)$ is evaluated by the mean of the $B$ posterior predictive distributions.

\Cref{fig:mp-illustraion-data} reflects a random sample of $F_\infty$ from the three specifications.  
The corresponding landscape of predictive uncertainty is depicted in \Cref{fig:mp-illustraion-uncertainty}. We can see that different specifications of future predictive distributions lead to different uncertainty quantification behaviors; for example, if we expect the future data to come from existing observations (as suggested by BB) or to be very similar to them (DP-MP with perturbed $H$), then there will be minimal uncertainty mostly concentrated around the decision boundary; however, if we expect almost ``zero'' prior knowledge (DP-MP with uniform $H$), then the uncertainty will be high except in the area with abundant evidence. 

It is critical to emphasize that there is no singular, correct specification of a Martingale posterior unless we have complete knowledge of the true data generative process. The key takeaway from this synthetic example is to illustrate that one can incorporate prior beliefs by defining a future predictive distribution, which will lead to a more informed representation of uncertainty.

\subsection{Experiment details for \Cref{fig:mixupdp-illustraion} and results for varying $\alpha$}
\label{app:subsec:illustraion-of-mixupmp}

In \Cref{fig:mixupdp-illustraion}, the synthetic data generation and the neural network architecture are the same as described in \Cref{app:subsec:illustration-of-mp}. We obtain their posterior samples $\{\theta^{(b)}\}_{b=1}^B$ according to \Cref{alg:ourmethod} with $B=10$ and $t_{mb}=n_{mb}$, and no standard data augmentation is used (i.e. $n_{mb}=n$). In each simulation, the network is trained from a random initialization with a learning rate of 0.5 for 10,000 epochs, and no mini-batch is used. 

\begin{figure}
    \centering
    \begin{subfigure}[b]{0.7\columnwidth}\includegraphics[width=\linewidth]{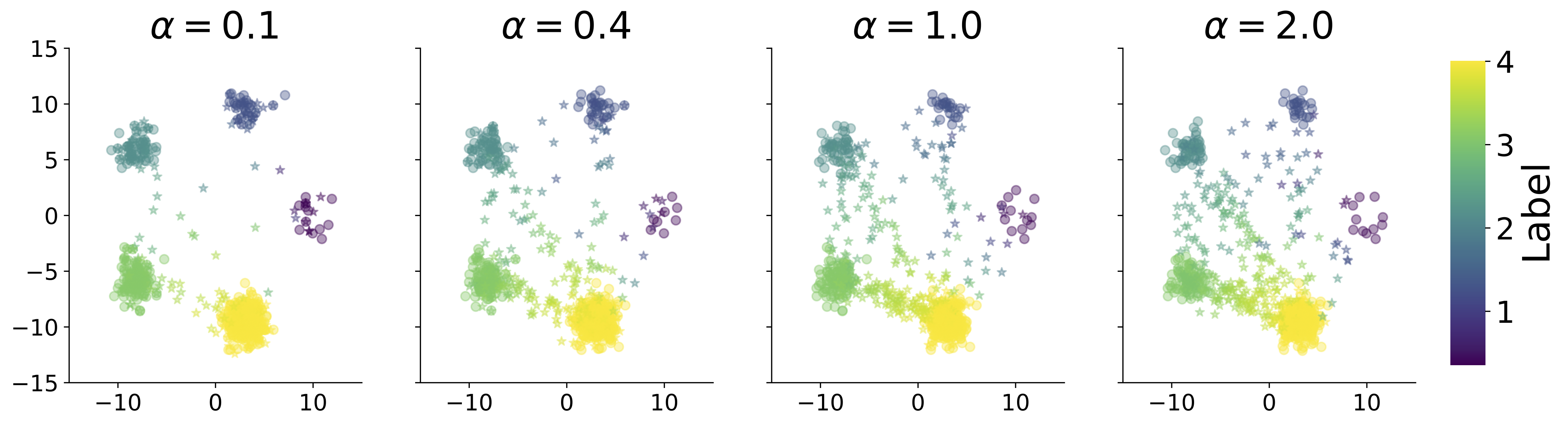}
    \caption{$\FMMP$ sample from \ourmethod{} with varying $\alpha \in [0.1,0.4,1.0,2.0]$,  $r=1$ and $\Haug_x:=\delta_x$. Dots represent samples from observations and crosses are samples from the base measure. The label space is extended to interval $[0,K-1]$.}
    \label{fig:mixupdp-illustraion-data-varying-alpha}
    \end{subfigure}
    \begin{subfigure}[b]{0.72\columnwidth}\includegraphics[width=\linewidth]{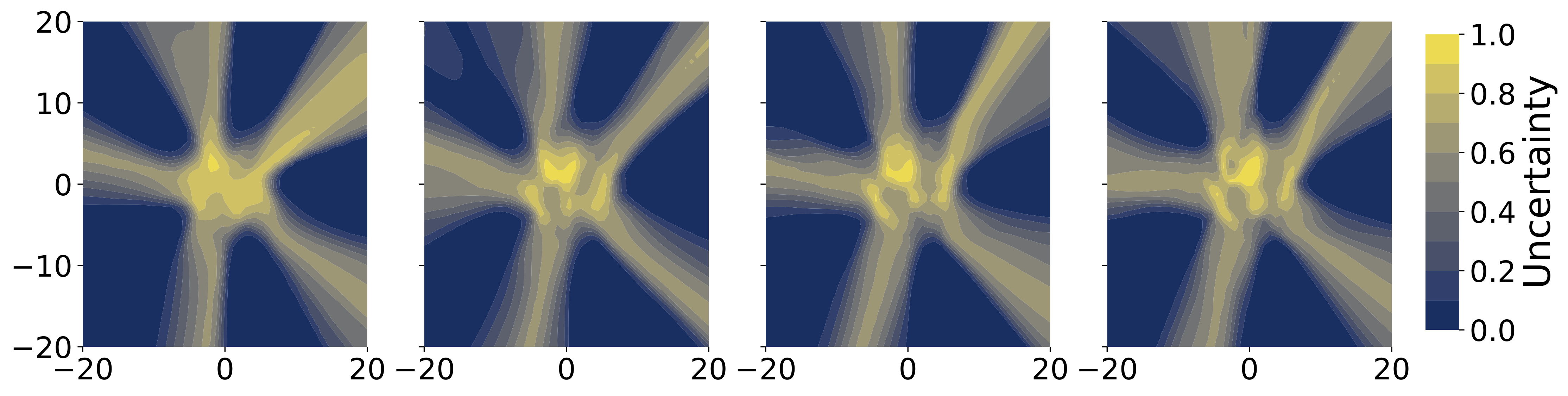}
    \caption{The corresponding predictive uncertainty landscape.}
    \label{fig:mixupdp-illustraion-uncertainty-varying-alpha}
    \end{subfigure}
    \caption{Illustration of \ourmethod{} on synthetic classification task $(K=5)$ with $r=1.0$ and varying $\alpha$. As  $\alpha$ increases, $\FMMP$ samples are more concentrated on the middle of different data pairs, inducing wider predictive uncertainty bands around the decision boundary.}
    \label{fig:mixupdp-illustraion-varying-alpha}
\end{figure}

Here we also include the illustration of \ourmethod{} with varying mixup parameter $\alpha$ in \Cref{fig:mixupdp-illustraion-varying-alpha}.

\section{Uncertainty calibration metrics}\label{app:uq_metrics}
For  uncertainty calibration metrics used in \Cref{sec:experiments}, we consider the expected calibration error \citep[ECE,][]{naeini2015obtaining}, over-confidence error (OE), and under-confidence error (UE). 

ECE  measures the difference between accuracy and confidence: 
\begin{align}
    \mbox {ECE} &= \sum_{m=1}^M \frac{|B_m|}{n} | \mbox{Acc} (B_m) - \mbox{Conf} (B_m) |
\end{align}
where for $m=1,\cdots, M$ 
\begin{align}
    \mbox{Acc} (B_m) &= \frac{1}{B_m} \sum_{x_i \in B_m} 
\mathbf{1} (\hat y_i = y_i), 
    \quad \mbox{Conf} (B_m) = \frac{1}{B_m} \sum_{x_i \in B_m} \hat p_i, 
\end{align}
each $B_m$ is the $m^\text{th}$ bin of samples whose prediction confidence falls into interval $(\frac{m-1}{M}, \frac{m}{M}]$,  $\hat y_i$ is the prediction and $\hat p_i$ is the prediction conference. 
We use $M=15$ throughout all experiments. 

We further inspect the over-confidence and under-confidence quantification:
\begin{align}
    \mbox{OE} &= \sum_{m=1}^M \frac{|B_m|}{n} \max \left\{\mbox{Conf} (B_m) - \mbox{Acc} (B_m) , 0\right\} \label{eq:oe} \\ 
        \mbox{UE} &= \sum_{m=1}^M \frac{|B_m|}{n} \max \left\{\mbox{Acc} (B_m) - \mbox{Conf} (B_m) , 0\right\} \label{eq:ue}
\end{align}
The study of OE and UE metrics are motivated by \citet{thulasidasan2019mixup}, where they propose another version of over-confidence metric that weights each summand in \Cref{eq:oe}  by the confidence score.

\section{Equivalency of BB and DE: Details and additional results}\label{app:equivalency}

\subsection{Experimental details}
To compare BB and DE, we looked at the behavior of ensembles of neural networks trained on separable data. Specifically, we looked at the MNIST dataset \citep[][ copyright the authors]{lecun1998mnist} and the Fashion MNIST (FMNIST) dataset \citep[][ MIT license]{xiao2017fashion}. Each ensemble contains 4 independently trained neural networks. We used a convolutional neural network with two convolutional layers (with kernel size 5; 6 and 16 hidden channels; ReLU activation; and max pooling) and two fully connected layers (with 120 and 84 latent dimensions and ReLU activation). 

For the ``random'' initialization results, each ensemble member was initialized using one of four randomly generated initializations. Initializations were paired between the two methods (e.g., the initialization of the first DE ensemble member is the same of that of the first BB DE ensemble member). Models were trained using stochastic gradient descent with a learning rate of $5e^{-4}$, for 1000 epochs beyond obtaining 100\% accuracy on the test set. These experiments were designed to explore the difference between BB and DE, given the same initialization. 

The ``DE'' initialization results were obtained using ensembles where each member was initialized to the output of one of the DE ``random'' initialization ensemble members. These results were designed to assess \Cref{prop:equiv}, which claims that a solution for DE is also a solution for BB. If this were not the case, we would expect the BB solution to diverge from the DE solution. Since the initialization already achieves 100\% training accuracy, models were trained using stochastic gradient descent with a learning rate of $5e^{-4}$, for 1000 epochs.

\Cref{tab:BBcomp_extended} (an extended version of \Cref{tab:BBcomp}, with additional metrics included) shows the performance of the resulting ensembles. We see that there is a small difference between the BB and DE results with random initialization, which we hypothesize is due to early training conditions. As expected, we see almost no difference when initialized to a pre-trained DE solution.

\begin{figure}[!t]
     \centering
     \begin{subfigure}[b]{\textwidth}
         \centering
         \includegraphics[width=\textwidth]{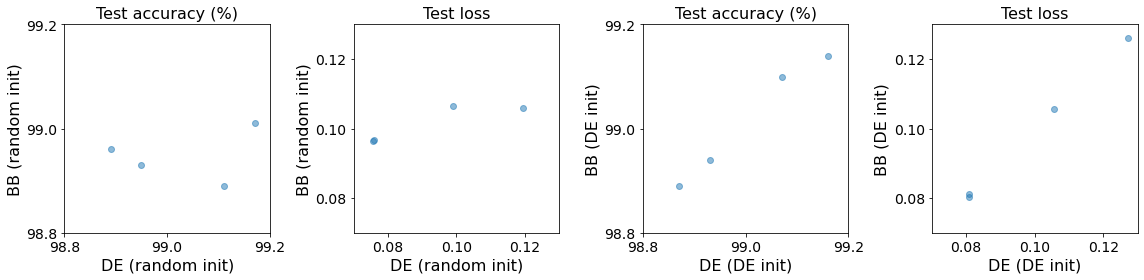}
         \caption{MNIST}
         \label{fig:seedcomp_MNIST_rand}
     \end{subfigure}
     \hfill
     \begin{subfigure}[b]{\textwidth}
         \centering
         \includegraphics[width=\textwidth]{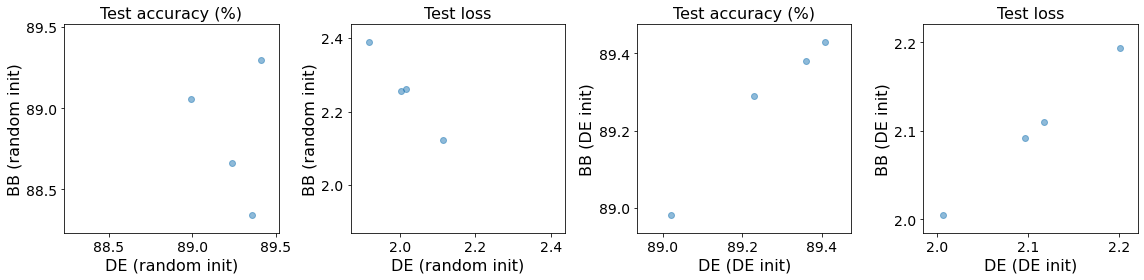}
         \caption{FashionMNIST }
         \label{fig:seedcomp_FMNIST_rand}
     \end{subfigure}
        \caption{Comparing individual ensemble members. Each point represents a pair of ensemble members from BB or DE with the same initialization (random or DE). }
        \label{fig:seed_comp}
\end{figure}

\Cref{fig:seed_comp} shows the test set accuracies and losses of each ensemble member in the BB and DE ensembles, pair-matched to have the same initialization. As in \Cref{tab:BBcomp_extended}, when randomly initialized, we see that the models obtain similar solutions, but not exactly the same (left two plots of each row). However, when pre-initialized to a DE solution, we see little deviation between the DE and BB individual ensemble members (right two plots of each row).


\begin{table*}[!th]
\centering 
\caption{Comparing Bayesian bootstrap (BB) with deep ensembles (DE). Models are either randomly initialized (same set of seeds for DE and BB), or initialized using a pretrained DE}
\label{tab:BBcomp_extended}
   \input{tables/bb_vs_de/bb_vs_de_full}
\end{table*}

\section{Empirical study of \ourmethod{}: experiment details and additional results}\label{app:extra_ourmethod_results}

\subsection{Implementation details.} 

We list implementation details of experiments in  \Cref{sec:ablation,sec:comparison,subsec:robustness}. 

For CIFAR10 and CIFAR100 datasets (under MIT license), we use the Wide Resnet  architecture with depth 28 and widen factor 10 for individual models \citep{zagoruyko2016wide}, and the implementation from \url{https://github.com/meliketoy/wide-resnet.pytorch} which is also under an MIT license. We follow the training details from the original paper: ``We use SGD with Nesterov momentum and cross-entropy loss. The
initial learning rate is set to 0.1, weight decay to 0.0005, dampening to 0, momentum to 0.9
and minibatch size to 128. Learning rate dropped by 0.2 at 60, 120 and 160 epochs
and we train for total 200 epochs.''

For FMNIST dataset, we use the Resnet18 architecture \citep{he2016deep}, which is under an MIT license, and the implementation from \url{https://github.com/kefth/fashion-mnist/tree/master}.  We additionally include the CNN architecture as an ablation (see results in \Cref{fig:ablation-fmnist-cnn}). Unless otherwise stated, the presented FMNIST results are obtained with Resnet18. The remaining training details are the same as the CIFAR experiment.

For implicit ensemble methods (namely MC Dropout, \ourmethod{}-MC, Mixup-MC, CAMixup-MC), a dropout rate of 0.3 is used. All remaining methods do not use dropout. 

CAMixup-MC requires using a validation set during training to evaluate the current ensemble's calibration behavior. We follow the practice from \citet{wen2020combining} to separate 5\% of data from the training set for validation purpose. 

For Laplace method, we use the implementation from the official codebase \url{https://github.com/aleximmer/Laplace}.  

For \ourmethod{} and its variants, we set the pseudo sample batch size $t_{mb}$ to the data batch size $n_{mb}$ (described in \Cref{alg:ourmethod}) throughout all experiments. Specifically, we used a batch size of 128. 

Lastly, for all ensemble methods (either explicit or implicit ensemble),  predictions are 
computed as the average of  predictions of individual models. 

\subsection{Ablation study on hyperparameters and data augmentations}
\label{app:subsec:ablation}

\begin{figure*}[!t]
    \centering
    \begin{subfigure}[b]{\textwidth}
    \includegraphics[width=\textwidth]{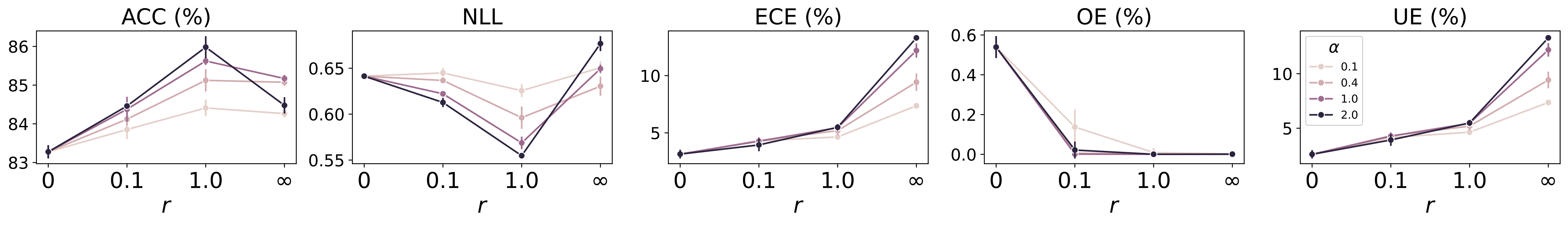}
    \caption{Performance on CIFAR100 dataset}
    \label{fig:ablation-cifar100}
    \end{subfigure}

    \begin{subfigure}[b]{\textwidth}
    \includegraphics[width=\textwidth]{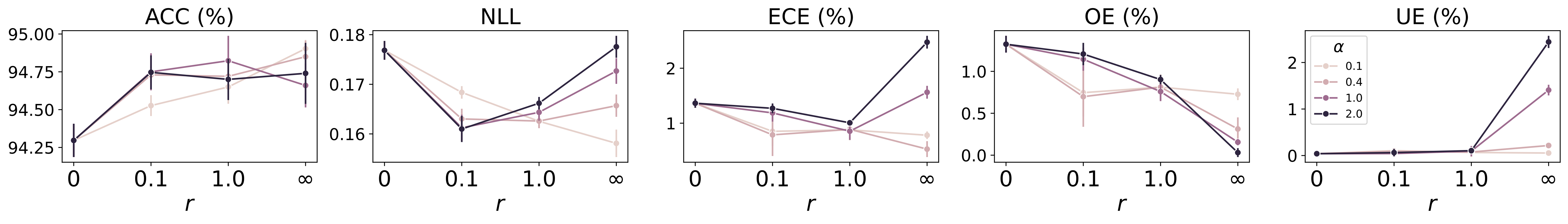}
    \caption{Performance on FMNIST dataset (ResNet18)}
    \label{fig:ablation-fmnist-resnet}
    \end{subfigure}

        \begin{subfigure}[b]{\textwidth}
    \includegraphics[width=\textwidth]{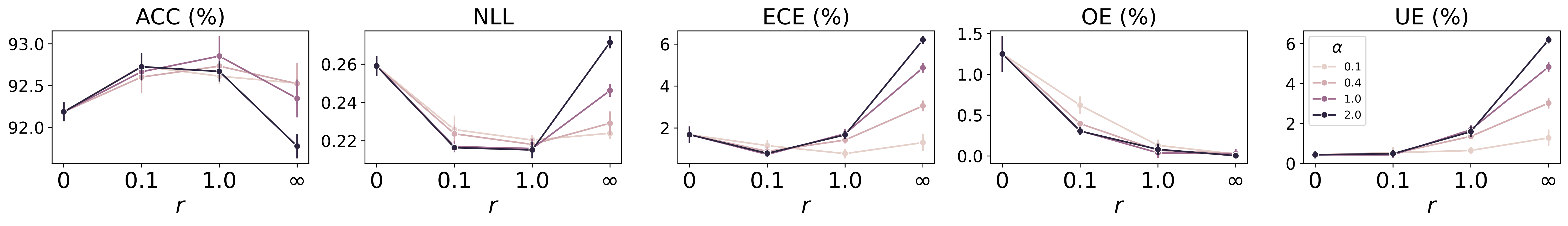}
    \caption{Performance on FMNIST dataset (CNN)}
    \label{fig:ablation-fmnist-cnn}
    \end{subfigure}
    
    \caption{\ourmethod{} ablation study: Impact of $\alpha$ and $r$ on test set performance on CIFAR100 (top row) and FMNIST (mid row: with ResNet18 and bottom row: with CNN). DE corresponds to \ourmethod{} with $r=0$; Mixup Ensemble corresponds to \ourmethod{} with $r=\infty$. Solid lines are average values computed over 3 random runs, with error bars denoting 2 standard errors.  The result for CIFAR10 is depicted in \Cref{fig:ablation-cifar10}.}
    \label{fig:ablation-cifar100-fmnist}
\end{figure*}

We include additional ablation results on varying ratio $r$ and varying mixup parameter $\alpha$ on the other two datasets in \Cref{fig:ablation-cifar100-fmnist}. 
We observe similar trends to those found in the result from the CIFAR10 dataset, as is shown in \Cref{fig:ablation-cifar10} and summarized in \Cref{sec:ablation}. However, the optimal choice of $r$ and $\alpha$ with respect to a specific metric may vary across datasets. In practice one can leave out an additional validation dataset to determine those values.



\subsection{Comparison to other methods}
\label{app:subsec:comparison}

To complement the study in \Cref{sec:comparison}, we summarize the in-distribution test results for all methods with  ablations on hyperparameters in \Cref{tab:comparison-all}. In additional to the methods listed in \Cref{sec:comparison}, we include \textit{\ourmethod{}-single}, which is a single  sample from \ourmethod{}  by running \Cref{alg:ourmethod} with $B=1$. 

We  observe that \ourmethod{} (including its variants \ourmethod{}-single and \ourmethod{}-MC) are the best or comparable to the best in terms of ACC and NLL in their corresponding group. 
The ECE performance of our method is either the best or close to best in each group, except that  MixupMP-MC ($r=0.4)$ on FMNIST dataset and CAMixup-MC ($\alpha=0.1,0.4$ or $1.0$) have better ECE.  Among all methods, \ourmethod{} via the explicit ensemble approach  strikes the best balance in predictive performance and calibration. 


\begin{table}[!t]

    \centering
    \caption{Full in-distribution test results on CIFAR10, CIFAR100 and FMNIST with standard errors included in the parenthesis. Bolded metrics are the best (within 2 standard errors) in each group of \{single model, explicit ensemble, implicit ensemble\}; $*$ indicates the best  among all methods.
     The blue shaded rows correspond to results of our method \ourmethod{} or its variants.}
    \label{tab:comparison-all}
    
\resizebox{\textwidth}{!}{%
\input{tables/comparison/test_ind_all_semTrue}
}
\end{table}

    

%

\subsection{Robustness to distribution shift}
\label{app:subsec:robustness}

We conduct a comprehensive study of \ourmethod{} in distribution shift / OOD (out-of-distribution) settings. For models trained on CIFAR10, we evaluate on CIFAR10-C \citep{hendrycks2018benchmarking} which contain 19 corruption datasets,  each with 5  intensity levels; 
For models trained on CIFAR100, we evaluate on CIFAR100-C \citep{hendrycks2018benchmarking}, which contain 21 corruption datasets, each with 5  intensity levels. Due to the large size of the test set, we only report results using a single run of each model.

\begin{figure}[!t]
    \centering
\caption{Performance under distribution shift on CIFAR100. We plot various metrics against the distribution shift intensity ranging from 0 to 5, where 0 indicates no shift.}
\label{fig:cifar100-robustness}

    \begin{subfigure}[b]{0.5\columnwidth}
\includegraphics[width=\linewidth]{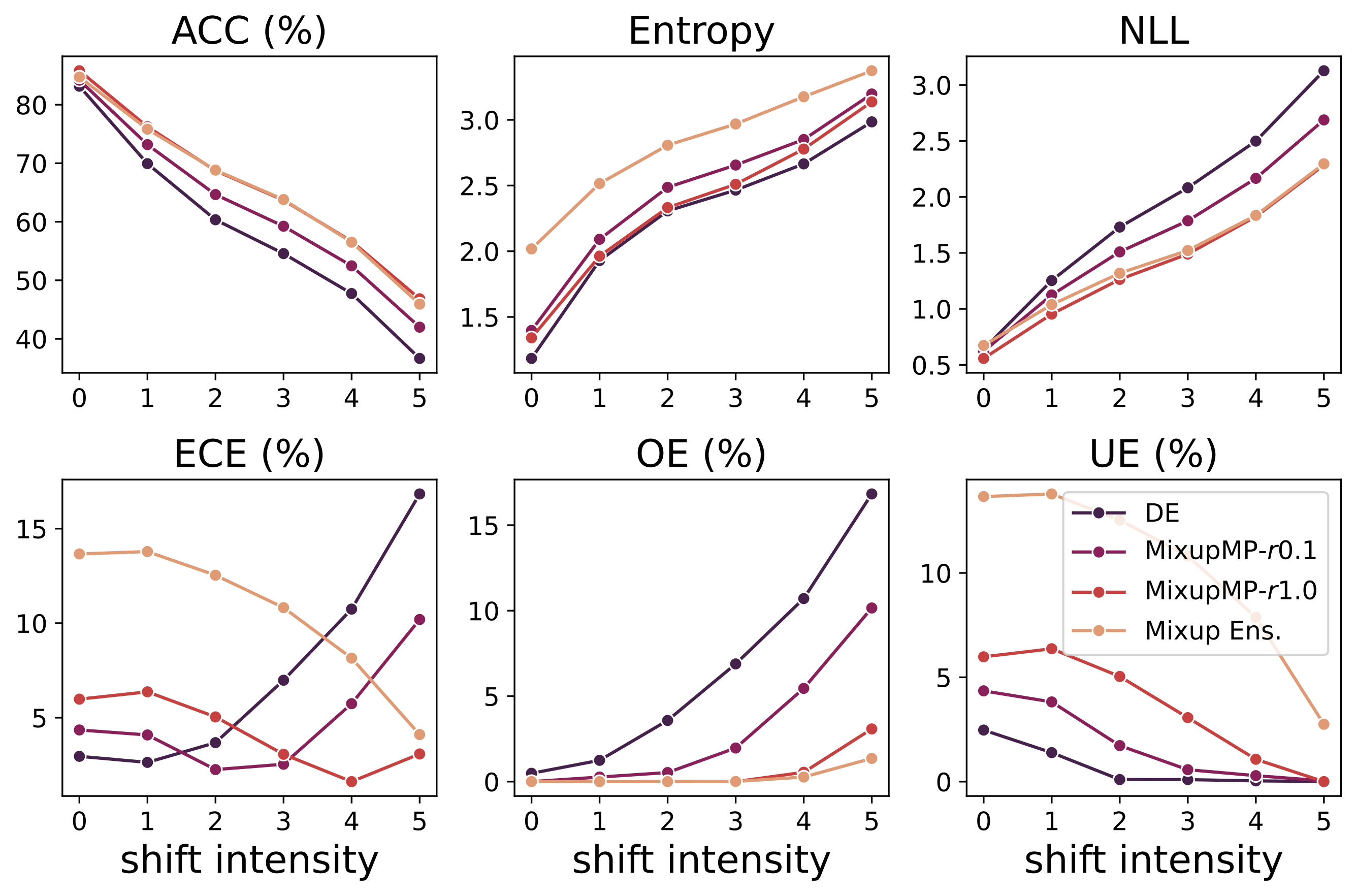}
\end{subfigure}

\end{figure}

For \ourmethod{} and its special cases DE and Mixup Ensemble, we  investigate their performance under different shift levels. The results for CIFAR10-C are summarized in \Cref{fig:cifar10-robustness}, and the results for CIFAR100-C are summarized in \Cref{fig:cifar100-robustness}. We observe similar trends in both plots; see detailed discussion in \Cref{subsec:robustness}. 

\paragraph{Full comparison.} In \Cref{tab:comparison-all-ood} we include a full comparison to all other methods listed in \Cref{sec:comparison,app:subsec:comparison}.

Across all methods, \ourmethod{} with large values of $r$ and $\alpha$ have superior OOD test performance in all metrics. Additionally, Mixup (single-model) and Mixup-MC also have favorable performance, especially in terms of ECE (they achieve the best ECE within their corresponding method group); however, Mixup Ensemble with a large $\alpha$ value (e.g. 1 or 2) can suffer from inflated ECE. 
Such results are expected, since with a large  $r$, \ourmethod{} (including Mixup Ensemble as a special case of $r=\infty$) has the prior belief that the future test data comes from a more uncertain domain away from the observations, which is the case in distribution shift settings.

While CAMixup-MC does well in in-distribution test set as shown in \Cref{tab:comparison-all}, its OOD test performance is worse than \ourmethod{} and Mixup Ensemble in most cases, especiallly in terms of ECE. This observation corroborates the findings in \citet{wen2020combining}. CAMixup  adjusts the calibration behavior using only in-distributoin validation dataset, explaining its performance does not generalize well to OOD settings. 

Lastly, we note that Laplace has inferior performance across all metrics, only slightly better than a single Neural Network (NN) and MixupMP-single with small $r$ and $\alpha$ (in which case is a very close model to NN). 

\begin{table}[!t]

    \centering
    \caption{Full out-of-distribution test results.  Bolded values indicate the best result within the group in \{ single model, explicit ensemble, implicit ensemble\}, and $*$ indicates the best among all methods. The blue shaded rows correspond to results of our method \ourmethod{} or its variants.} 
    \label{tab:comparison-all-ood}
    
\resizebox{0.8\textwidth}{!}{%
\input{tables/robustness-to-ood/test_ood_all_no_sem}

}
\end{table}

\section{Time and space complexity of methods described in the paper}\label{app:complexity}
On a per-epoch basis, the time and memory complexity of the MP methods described in this paper are of the same order of magnitude as the corresponding DE method. In the case of \ourmethod{}, each minibatch is of length $t_{mb} + n_{mb}$, increasing the time and memory complexity by a constant factor. For BB, we have the same minibatch sizes as in DE; the additional cost of adding in weights is negligible.

\clearpage

\section*{Checklist}



 \begin{enumerate}

 \item For all models and algorithms presented, check if you include:
 \begin{enumerate}
   \item A clear description of the mathematical setting, assumptions, algorithm, and/or model. [Yes] See \Cref{sec:mixupmp}
   \item An analysis of the properties and complexity (time, space, sample size) of any algorithm. [Yes] See \Cref{app:complexity}
   \item (Optional) Anonymized source code, with specification of all dependencies, including external libraries. [Yes] \url{https://github.com/apple/ml-MixupMP}.
 \end{enumerate}

 \item For any theoretical claim, check if you include:
 \begin{enumerate}
   \item Statements of the full set of assumptions of all theoretical results. [Yes] See \Cref{appendix-sec:proof}.
   \item Complete proofs of all theoretical results. [Yes] See \Cref{appendix-sec:proof}
   \item Clear explanations of any assumptions. [Yes] See \Cref{appendix-sec:proof}.     
 \end{enumerate}

 \item For all figures and tables that present empirical results, check if you include:
 \begin{enumerate}
   \item The code, data, and instructions needed to reproduce the main experimental results (either in the supplemental material or as a URL). [Yes] \url{https://github.com/apple/ml-MixupMP}.
   \item All the training details (e.g., data splits, hyperparameters, how they were chosen). [Yes] See relevant sections of the appendix
         \item A clear definition of the specific measure or statistics and error bars (e.g., with respect to the random seed after running experiments multiple times). [Yes] 
         The metric definitions and the number of runs are  described in   the experiment section and appendix.  The error bars in  figures denote 2 standard errors. 
         \item A description of the computing infrastructure used. (e.g., type of GPUs, internal cluster, or cloud provider). [Yes] Experiments were carried out using Apple internal clusters. We do not include full compute information for privacy reasons; however none of our results require a specific infrastructure. 
 \end{enumerate}

 \item If you are using existing assets (e.g., code, data, models) or curating/releasing new assets, check if you include:
 \begin{enumerate}
   \item Citations of the creator If your work uses existing assets. [Yes] See \Cref{app:equivalency} and \Cref{app:extra_ourmethod_results}.
   \item The license information of the assets, if applicable. [Yes] See \Cref{app:equivalency} and \Cref{app:extra_ourmethod_results}.
   \item New assets either in the supplemental material or as a URL, if applicable. [No]
   \item Information about consent from data providers/curators. [Not Applicable]
   \item Discussion of sensible content if applicable, e.g., personally identifiable information or offensive content. [Not Applicable]
 \end{enumerate}

 \item If you used crowdsourcing or conducted research with human subjects, check if you include:
 \begin{enumerate}
   \item The full text of instructions given to participants and screenshots. [Not Applicable]
   \item Descriptions of potential participant risks, with links to Institutional Review Board (IRB) approvals if applicable. [Not Applicable]
   \item The estimated hourly wage paid to participants and the total amount spent on participant compensation. [Not Applicable]
 \end{enumerate}

 \end{enumerate}


%% file: tables/bb_vs_de/bb_vs_de_full.tex
 \begin{tabular}{cccccccc}
    \hline 
    Dataset & Method & Init. & ACC (\%) & ECE (\%) & OE (\%) &  UE (\%) & NLL\\ \hline 
MNIST & DE & random & 99.33 & 0.4092 & 0.2224& 0.1868 & 0.035111\\
MNIST & BB & random & 99.17 & 0.2357 & 0.2133& 0.0224 & 0.037333\\ \hline 
MNIST & DE & DE & 99.33 & 0.3997 & 0.2185& 0.1812 & 0.036600\\ 
MNIST & BB & DE & 99.33 & 0.4079 & 0.2195& 0.1884 & 0.036653\\ \hline 
FMNIST & DE & random & 91.52 & 2.3240 & 2.1323& 0.1917 & 0.571902\\
FMNIST & BB & random & 91.21 & 2.0132 & 1.8851& 0.1281 & 0.620820\\ \hline 
FMNIST & DE & DE & 91.57 & 2.4219 & 2.1655& 0.2564 & 0.592023\\ 
FMNIST & BB & DE & 91.55 & 2.3866 & 2.1616& 0.2249 & 0.591654\\ \hline 
\end{tabular}

%% file: tables/comparison/test_ind_all_semTrue.tex
\begin{tabular}{llll|lll|lll} \hline 
\multicolumn{1}{l|}{Dataset}& \multicolumn{3}{l|}{CIFAR10}& \multicolumn{3}{l|}{CIFAR100}& \multicolumn{3}{l}{FMNIST} \\ \hline 
\multicolumn{1}{l|}{Metric} & ACC (\%)  & NLL  & ECE (\%)  & ACC (\%)  & NLL  & ECE (\%)  & ACC (\%)  & NLL  & ECE (\%) \\ \hline 
\multicolumn{1}{l|}{\textbf{Single model}}  &  &  &  &  &  &  &  &  & \\ 
\multicolumn{1}{l|}{NN} &$96.12 (0.04)$ &$0.1473 (0.0012)$ &$2.02 (0.04)$ &$80.82 (0.08)$ &$0.7816 (0.0026)$ &$4.78 (0.08)$ &$93.72 (0.08)$ &$0.2181 (0.0024)$ &$2.89 (0.06)$\\ 
\rowcolor[HTML]{E6ECFC} \multicolumn{1}{l|}{MixupMP-single (r=0.1, $\alpha$=0.1)} &$96.35 (0.04)$ &$0.1356 (0.0010)$ &$1.26 (0.04)$ &$81.30 (0.06)$ &$0.7712 (0.0021)$ &$3.66 (0.11)$ &$94.09 (0.05)$ &$0.1959 (0.0010)$ &$1.95 (0.11)$\\ 
\rowcolor[HTML]{E6ECFC} \multicolumn{1}{l|}{MixupMP-single (r=0.1, $\alpha$=0.4)} &$96.51 (0.03)$ &$0.1302 (0.0009)$ &$0.95 (0.03)$ &$81.85 (0.08)$ &$0.7535 (0.0023)$ &$3.54 (0.06)$ &$94.23 (0.03)$ &$0.1914 (0.0012)$ &$2.01 (0.12)$\\ 
\rowcolor[HTML]{E6ECFC} \multicolumn{1}{l|}{MixupMP-single (r=0.1, $\alpha$=1.0)} &$96.65 (0.04)$ &$0.1258 (0.0012)$ &$0.91 (0.05)$ &$82.22 (0.06)$ &$0.7322 (0.0020)$ &$3.23 (0.08)$ &$94.31 (0.05)$ &$0.1890 (0.0011)$ &$2.29 (0.03)$\\ 
\rowcolor[HTML]{E6ECFC} \multicolumn{1}{l|}{MixupMP-single (r=0.1, $\alpha$=2.0)} &$96.62 (0.03)$ &$0.1260 (0.0008)$ &$1.05 (0.02)$ &$82.47 (0.06)$ &$0.7201 (0.0020)$ &$3.24 (0.11)$ &$94.31 (0.03)$ &$0.1895 (0.0009)$ &$2.35 (0.05)$\\ 
\rowcolor[HTML]{E6ECFC} \multicolumn{1}{l|}{MixupMP-single (r=1.0, $\alpha$=0.1)} &$96.57 (0.03)$ &$0.1288 (0.0010)$ &$\mathbf{0.46 (0.04)}$ &$82.09 (0.05)$ &$0.7352 (0.0025)$ &$2.72 (0.09)$ &$94.21 (0.04)$ &$0.1870 (0.0010)$ &$1.92 (0.05)$\\ 
\rowcolor[HTML]{E6ECFC} \multicolumn{1}{l|}{MixupMP-single (r=1.0, $\alpha$=0.4)} &$96.87 (0.04)$ &$0.1202 (0.0011)$ &$0.68 (0.04)$ &$82.86 (0.07)$ &$0.7046 (0.0022)$ &$1.66 (0.08)$ &$94.25 (0.04)$ &$0.1893 (0.0010)$ &$1.96 (0.04)$\\ 
\rowcolor[HTML]{E6ECFC} \multicolumn{1}{l|}{MixupMP-single (r=1.0, $\alpha$=1.0)} &$97.06 (0.03)$ &$0.1130 (0.0008)$ &$0.77 (0.04)$ &$83.27 (0.05)$ &$0.6769 (0.0024)$ &$\mathbf{1.28 (0.09)}^*$ &$94.27 (0.04)$ &$0.1913 (0.0011)$ &$2.03 (0.03)$\\ 
\rowcolor[HTML]{E6ECFC} \multicolumn{1}{l|}{MixupMP-single (r=1.0, $\alpha$=2.0)} &$\mathbf{97.27 (0.03)}$ &$\mathbf{0.1055 (0.0009)}$ &$0.62 (0.05)$ &$\mathbf{83.56 (0.05)}$ &$\mathbf{0.6582 (0.0012)}$ &$\mathbf{1.13 (0.06)}^*$ &$94.19 (0.04)$ &$0.1938 (0.0009)$ &$2.10 (0.04)$\\ 
\multicolumn{1}{l|}{Mixup ($\alpha$=0.1)} &$96.40 (0.03)$ &$0.1354 (0.0010)$ &$0.82 (0.05)$ &$81.82 (0.06)$ &$0.7412 (0.0029)$ &$3.30 (0.09)$ &$\mathbf{94.48 (0.03)}$ &$\mathbf{0.1833 (0.0010)}$ &$1.79 (0.03)$\\ 
\multicolumn{1}{l|}{Mixup ($\alpha$=0.4)} &$96.74 (0.03)$ &$0.1378 (0.0011)$ &$2.68 (0.05)$ &$82.69 (0.05)$ &$0.7156 (0.0030)$ &$5.11 (0.20)$ &$94.33 (0.03)$ &$0.1902 (0.0008)$ &$1.35 (0.08)$\\ 
\multicolumn{1}{l|}{Mixup ($\alpha$=1.0)} &$96.88 (0.04)$ &$0.1566 (0.0017)$ &$5.36 (0.18)$ &$82.59 (0.07)$ &$0.7339 (0.0021)$ &$7.61 (0.22)$ &$94.20 (0.04)$ &$0.1933 (0.0009)$ &$\mathbf{0.81 (0.03)}$\\ 
\multicolumn{1}{l|}{Mixup ($\alpha$=2.0)} &$96.88 (0.03)$ &$0.1854 (0.0016)$ &$8.59 (0.14)$ &$81.83 (0.09)$ &$0.7614 (0.0028)$ &$8.54 (0.19)$ &$94.15 (0.05)$ &$0.1948 (0.0008)$ &$1.46 (0.05)$\\ 
\multicolumn{1}{l|}{Laplace} &$96.04 (0.05)$ &$0.1344 (0.0010)$ &$0.80 (0.04)$ &$80.95 (0.18)$ &$1.0107 (0.0124)$ &$22.96 (1.02)$ &$89.67 (0.10)$ &$0.4088 (0.0045)$ &$5.89 (0.05)$\\ 
\hline 
\multicolumn{1}{l|}{\textbf{Explicit ensemble (B=4)}}  &  &  &  &  &  &  &  &  & \\ 
\multicolumn{1}{l|}{DE} &$96.83 (0.02)$ &$0.1090 (0.0007)$ &$0.78 (0.04)$ &$83.28 (0.08)$ &$0.6413 (0.0003)$ &$\mathbf{3.14 (0.15)}$ &$94.30 (0.05)$ &$0.1768 (0.0009)$ &$1.36 (0.03)$\\ 
\rowcolor[HTML]{E6ECFC} \multicolumn{1}{l|}{MixupMP (r=0.1, $\alpha$=0.1)} &$97.01 (0.05)$ &$0.1042 (0.0007)$ &$\mathbf{0.40 (0.07)}^*$ &$83.85 (0.11)$ &$0.6449 (0.0023)$ &$4.29 (0.03)$ &$94.53 (0.03)$ &$0.1684 (0.0006)$ &$0.85 (0.13)$\\ 
\rowcolor[HTML]{E6ECFC} \multicolumn{1}{l|}{MixupMP (r=1.0, $\alpha$=0.1)} &$97.17 (0.02)$ &$0.1014 (0.0011)$ &$1.13 (0.07)$ &$84.41 (0.10)$ &$0.6256 (0.0030)$ &$4.64 (0.04)$ &$94.65 (0.05)$ &$0.1625 (0.0006)$ &$0.88 (0.09)$\\ 
\rowcolor[HTML]{E6ECFC} \multicolumn{1}{l|}{MixupMP (r=0.1, $\alpha$=0.4)} &$97.04 (0.03)$ &$0.1009 (0.0007)$ &$0.55 (0.03)$ &$84.12 (0.07)$ &$0.6367 (0.0008)$ &$4.31 (0.09)$ &$94.73 (0.03)$ &$0.1630 (0.0009)$ &$\mathbf{0.79 (0.19)}^*$\\ 
\rowcolor[HTML]{E6ECFC} \multicolumn{1}{l|}{MixupMP (r=1.0, $\alpha$=0.4)} &$97.39 (0.02)$ &$0.0951 (0.0015)$ &$1.67 (0.03)$ &$85.12 (0.13)$ &$0.5962 (0.0056)$ &$5.19 (0.16)$ &$94.72 (0.02)$ &$0.1626 (0.0006)$ &$0.89 (0.03)$\\ 
\rowcolor[HTML]{E6ECFC} \multicolumn{1}{l|}{MixupMP (r=0.1, $\alpha$=1.0)} &$97.13 (0.04)$ &$0.0966 (0.0016)$ &$0.59 (0.02)$ &$84.38 (0.15)$ &$0.6222 (0.0007)$ &$4.27 (0.14)$ &$94.75 (0.06)$ &$0.1612 (0.0012)$ &$1.19 (0.07)$\\ 
\rowcolor[HTML]{E6ECFC} \multicolumn{1}{l|}{MixupMP (r=1.0, $\alpha$=1.0)} &$97.57 (0.05)$ &$0.0918 (0.0004)$ &$1.78 (0.09)$ &$85.62 (0.04)$ &$0.5687 (0.0030)$ &$5.45 (0.12)$ &$\mathbf{94.82 (0.08)}^*$ &$0.1644 (0.0008)$ &$0.85 (0.07)$\\ 
\rowcolor[HTML]{E6ECFC} \multicolumn{1}{l|}{MixupMP (r=0.1, $\alpha$=2.0)} &$97.13 (0.03)$ &$0.0969 (0.0004)$ &$\mathbf{0.46 (0.03)}$ &$84.46 (0.03)$ &$0.6127 (0.0021)$ &$3.95 (0.24)$ &$94.75 (0.05)$ &$0.1610 (0.0012)$ &$1.27 (0.04)$\\ 
\rowcolor[HTML]{E6ECFC} \multicolumn{1}{l|}{MixupMP (r=1.0, $\alpha$=2.0)} &$\mathbf{97.77 (0.02)}^*$ &$\mathbf{0.0845 (0.0010)}^*$ &$1.67 (0.08)$ &$\mathbf{85.98 (0.13)}^*$ &$\mathbf{0.5548 (0.0011)}^*$ &$5.49 (0.10)$ &$94.70 (0.07)$ &$0.1662 (0.0005)$ &$1.01 (0.00)$\\ 
\multicolumn{1}{l|}{Mixup Ensemble ($\alpha$=0.1)} &$97.03 (0.02)$ &$0.1099 (0.0005)$ &$1.91 (0.05)$ &$84.26 (0.04)$ &$0.6506 (0.0029)$ &$7.36 (0.06)$ &$\mathbf{94.90 (0.03)}^*$ &$\mathbf{0.1581 (0.0013)}^*$ &$0.78 (0.03)$\\ 
\multicolumn{1}{l|}{Mixup Ensemble ($\alpha$=0.4)} &$97.32 (0.09)$ &$0.1178 (0.0009)$ &$3.86 (0.02)$ &$85.07 (0.04)$ &$0.6304 (0.0048)$ &$9.43 (0.35)$ &$\mathbf{94.85 (0.05)}^*$ &$0.1657 (0.0010)$ &$\mathbf{0.53 (0.06)}^*$\\ 
\multicolumn{1}{l|}{Mixup Ensemble ($\alpha$=1.0)} &$97.47 (0.01)$ &$0.1395 (0.0018)$ &$6.55 (0.12)$ &$85.17 (0.04)$ &$0.6495 (0.0018)$ &$12.19 (0.27)$ &$94.66 (0.07)$ &$0.1727 (0.0012)$ &$1.56 (0.05)$\\ 
\multicolumn{1}{l|}{Mixup Ensemble ($\alpha$=2.0)} &$97.55 (0.02)$ &$0.1717 (0.0022)$ &$9.71 (0.18)$ &$84.48 (0.10)$ &$0.6768 (0.0036)$ &$13.30 (0.04)$ &$\mathbf{94.74 (0.10)}^*$ &$0.1776 (0.0010)$ &$2.47 (0.05)$\\ 
\hline 
\multicolumn{1}{l|}{\textbf{Implicit ensemble (B=20)}}  &  &  &  &  &  &  &  &  & \\ 
\multicolumn{1}{l|}{MC Dropout} &$96.16 (0.05)$ &$0.1315 (0.0011)$ &$1.46 (0.04)$ &$80.83 (0.25)$ &$0.7451 (0.0045)$ &$3.69 (0.07)$ &$94.41 (0.04)$ &$0.1806 (0.0004)$ &$1.90 (0.05)$\\ 
\rowcolor[HTML]{E6ECFC} \multicolumn{1}{l|}{MixupMP-MC (r=0.1, $\alpha$=0.1)} &$96.49 (0.10)$ &$0.1204 (0.0035)$ &$0.72 (0.11)$ &$81.78 (0.16)$ &$0.7215 (0.0012)$ &$3.17 (0.05)$ &$94.57 (0.06)$ &$0.1711 (0.0006)$ &$1.42 (0.06)$\\ 
\rowcolor[HTML]{E6ECFC} \multicolumn{1}{l|}{MixupMP-MC (r=1.0, $\alpha$=0.1)} &$96.75 (0.02)$ &$0.1163 (0.0014)$ &$0.57 (0.05)$ &$82.08 (0.22)$ &$0.7102 (0.0035)$ &$2.76 (0.33)$ &$\mathbf{94.76 (0.08)}^*$ &$0.1646 (0.0005)$ &$1.27 (0.06)$\\ 
\rowcolor[HTML]{E6ECFC} \multicolumn{1}{l|}{MixupMP-MC (r=0.1, $\alpha$=0.4)} &$96.62 (0.04)$ &$0.1180 (0.0010)$ &$\mathbf{0.46 (0.08)}^*$ &$81.80 (0.10)$ &$0.7276 (0.0019)$ &$3.07 (0.04)$ &$\mathbf{94.65 (0.16)}^*$ &$0.1674 (0.0025)$ &$1.44 (0.10)$\\ 
\rowcolor[HTML]{E6ECFC} \multicolumn{1}{l|}{MixupMP-MC (r=1.0, $\alpha$=0.4)} &$97.09 (0.03)$ &$0.1103 (0.0011)$ &$1.12 (0.04)$ &$83.32 (0.11)$ &$0.6825 (0.0057)$ &$4.92 (0.13)$ &$\mathbf{94.81 (0.05)}^*$ &$\mathbf{0.1619 (0.0005)}$ &$0.98 (0.04)$\\ 
\rowcolor[HTML]{E6ECFC} \multicolumn{1}{l|}{MixupMP-MC (r=0.1, $\alpha$=1.0)} &$96.77 (0.01)$ &$0.1132 (0.0018)$ &$\mathbf{0.35 (0.06)}^*$ &$82.50 (0.07)$ &$0.7067 (0.0033)$ &$3.55 (0.07)$ &$\mathbf{94.78 (0.05)}$ &$0.1648 (0.0008)$ &$1.38 (0.06)$\\ 
\rowcolor[HTML]{E6ECFC} \multicolumn{1}{l|}{MixupMP-MC (r=1.0, $\alpha$=1.0)} &$\mathbf{97.15 (0.08)}$ &$0.1063 (0.0013)$ &$1.20 (0.10)$ &$\mathbf{83.55 (0.03)}$ &$0.6520 (0.0020)$ &$4.23 (0.26)$ &$\mathbf{94.86 (0.05)}^*$ &$\mathbf{0.1624 (0.0012)}$ &$0.90 (0.02)$\\ 
\rowcolor[HTML]{E6ECFC} \multicolumn{1}{l|}{MixupMP-MC (r=0.1, $\alpha$=2.0)} &$96.58 (0.05)$ &$0.1145 (0.0029)$ &$0.42 (0.02)$ &$82.42 (0.12)$ &$0.7079 (0.0045)$ &$3.39 (0.10)$ &$94.72 (0.07)$ &$0.1651 (0.0015)$ &$1.46 (0.02)$\\ 
\rowcolor[HTML]{E6ECFC} \multicolumn{1}{l|}{MixupMP-MC (r=1.0, $\alpha$=2.0)} &$\mathbf{97.27 (0.04)}$ &$\mathbf{0.1005 (0.0006)}$ &$1.21 (0.08)$ &$\mathbf{83.58 (0.05)}$ &$\mathbf{0.6419 (0.0043)}$ &$4.21 (0.10)$ &$\mathbf{94.81 (0.06)}^*$ &$\mathbf{0.1619 (0.0007)}$ &$0.97 (0.04)$\\ 
\multicolumn{1}{l|}{Mixup-MC ($\alpha$=0.1)} &$96.55 (0.06)$ &$0.1272 (0.0013)$ &$1.43 (0.07)$ &$81.77 (0.20)$ &$0.7263 (0.0025)$ &$3.46 (0.38)$ &$\mathbf{94.84 (0.05)}^*$ &$0.1649 (0.0007)$ &$1.00 (0.06)$\\ 
\multicolumn{1}{l|}{Mixup-MC ($\alpha$=0.4)} &$96.68 (0.03)$ &$0.1394 (0.0013)$ &$3.38 (0.14)$ &$82.52 (0.06)$ &$0.7235 (0.0029)$ &$7.49 (0.21)$ &$\mathbf{94.86 (0.06)}^*$ &$0.1656 (0.0008)$ &$\mathbf{0.55 (0.05)}^*$\\ 
\multicolumn{1}{l|}{Mixup-MC ($\alpha$=1.0)} &$96.96 (0.07)$ &$0.1613 (0.0012)$ &$6.71 (0.27)$ &$82.26 (0.05)$ &$0.7331 (0.0087)$ &$8.98 (0.76)$ &$94.73 (0.03)$ &$0.1748 (0.0007)$ &$1.81 (0.04)$\\ 
\multicolumn{1}{l|}{Mixup-MC ($\alpha$=2.0)} &$96.80 (0.02)$ &$0.2084 (0.0045)$ &$10.98 (0.26)$ &$81.75 (0.08)$ &$0.7616 (0.0046)$ &$10.71 (0.20)$ &$94.69 (0.03)$ &$0.1796 (0.0007)$ &$2.84 (0.10)$\\ 
\multicolumn{1}{l|}{CAMixup-MC ($\alpha$=0.1)} &$95.82 (0.05)$ &$0.1529 (0.0083)$ &$\mathbf{1.86 (0.85)}^*$ &$79.83 (0.44)$ &$0.7942 (0.0250)$ &$\mathbf{2.44 (0.40)}$ &$94.13 (0.06)$ &$0.1808 (0.0025)$ &$1.13 (0.04)$\\ 
\multicolumn{1}{l|}{CAMixup-MC ($\alpha$=0.4)} &$95.72 (0.05)$ &$0.1571 (0.0080)$ &$2.52 (0.68)$ &$80.49 (0.14)$ &$0.7647 (0.0071)$ &$\mathbf{1.97 (0.25)}$ &$94.19 (0.04)$ &$0.1817 (0.0019)$ &$0.86 (0.14)$\\ 
\multicolumn{1}{l|}{CAMixup-MC ($\alpha$=1.0)} &$95.94 (0.03)$ &$0.1466 (0.0076)$ &$1.63 (0.62)$ &$80.21 (0.19)$ &$0.7739 (0.0122)$ &$\mathbf{2.05 (0.20)}$ &$94.31 (0.05)$ &$0.1788 (0.0012)$ &$0.97 (0.09)$\\ 
\multicolumn{1}{l|}{CAMixup-MC ($\alpha$=2.0)} &$96.11 (0.10)$ &$0.1365 (0.0070)$ &$1.21 (0.08)$ &$80.19 (0.11)$ &$0.7780 (0.0061)$ &$2.14 (0.04)$ &$94.27 (0.04)$ &$0.1818 (0.0020)$ &$1.04 (0.14)$\\ 
\hline 
 \end{tabular}

%% file: tables/robustness-to-ood/test_ood_all_no_sem.tex
\begin{tabular}{llll|lll} \hline 
\multicolumn{1}{l|}{Dataset}& \multicolumn{3}{l|}{CIFAR10-C}& \multicolumn{3}{l}{CIFAR100-C} \\ \hline 
\multicolumn{1}{l|}{Metric} & ACC (\%)  & NLL  & ECE (\%)  & ACC (\%)  & NLL  & ECE (\%) \\ \hline 
\multicolumn{1}{l|}{\textbf{Single model}}  &  &  &  &  &  & \\ 
\multicolumn{1}{l|}{NN} &$75.50$ &$1.1287$ &$16.03$ &$50.81$ &$2.4562$ &$15.93$\\ 
\rowcolor[HTML]{E6ECFC} \multicolumn{1}{l|}{MixupMP-single (r=0.1, $\alpha$=0.1)} &$77.21$ &$0.9422$ &$12.91$ &$53.31$ &$2.2401$ &$12.60$\\ 
\rowcolor[HTML]{E6ECFC} \multicolumn{1}{l|}{MixupMP-single (r=0.1, $\alpha$=0.4)} &$78.78$ &$0.8188$ &$10.55$ &$54.03$ &$2.1875$ &$10.96$\\ 
\rowcolor[HTML]{E6ECFC} \multicolumn{1}{l|}{MixupMP-single (r=0.1, $\alpha$=1.0)} &$79.67$ &$0.7787$ &$9.95$ &$55.41$ &$2.0825$ &$9.29$\\ 
\rowcolor[HTML]{E6ECFC} \multicolumn{1}{l|}{MixupMP-single (r=0.1, $\alpha$=2.0)} &$80.38$ &$0.7479$ &$9.29$ &$55.60$ &$2.0860$ &$9.94$\\ 
\rowcolor[HTML]{E6ECFC} \multicolumn{1}{l|}{MixupMP-single (r=1.0, $\alpha$=0.1)} &$79.96$ &$0.7725$ &$9.33$ &$54.88$ &$2.0636$ &$8.24$\\ 
\rowcolor[HTML]{E6ECFC} \multicolumn{1}{l|}{MixupMP-single (r=1.0, $\alpha$=0.4)} &$81.42$ &$0.7134$ &$7.97$ &$57.31$ &$1.9141$ &$6.57$\\ 
\rowcolor[HTML]{E6ECFC} \multicolumn{1}{l|}{MixupMP-single (r=1.0, $\alpha$=1.0)} &$81.95$ &$0.6887$ &$7.48$ &$58.49$ &$1.8216$ &$5.79$\\ 
\rowcolor[HTML]{E6ECFC} \multicolumn{1}{l|}{MixupMP-single (r=1.0, $\alpha$=2.0)} &$\mathbf{83.02}$ &$0.6324$ &$6.06$ &$\mathbf{58.94}$ &$\mathbf{1.7755}$ &$5.67$\\ 
\multicolumn{1}{l|}{Mixup ($\alpha$=0.1)} &$80.77$ &$0.7046$ &$6.96$ &$54.65$ &$2.0589$ &$7.72$\\ 
\multicolumn{1}{l|}{Mixup ($\alpha$=0.4)} &$81.66$ &$0.6625$ &$4.11$ &$57.24$ &$1.8749$ &$4.78$\\ 
\multicolumn{1}{l|}{Mixup ($\alpha$=1.0)} &$82.04$ &$\mathbf{0.6286}$ &$\mathbf{2.34}$ &$57.52$ &$1.8255$ &$\mathbf{1.63}$\\ 
\multicolumn{1}{l|}{Mixup ($\alpha$=2.0)} &$80.75$ &$0.6786$ &$4.70$ &$57.99$ &$1.7932$ &$2.16$\\ 
\multicolumn{1}{l|}{Laplace} &$75.87$ &$0.8933$ &$11.98$ &$51.26$ &$2.3194$ &$15.53$\\ 
\hline 
\multicolumn{1}{l|}{\textbf{Explicit ensemble (B=4)}}  &  &  &  &  &  & \\ 
\multicolumn{1}{l|}{DE} &$77.03$ &$0.8968$ &$10.29$ &$53.84$ &$2.1383$ &$7.65$\\ 
\rowcolor[HTML]{E6ECFC} \multicolumn{1}{l|}{MixupMP (r=0.1, $\alpha$=0.1)} &$78.79$ &$0.7410$ &$7.01$ &$56.35$ &$1.9735$ &$4.78$\\ 
\rowcolor[HTML]{E6ECFC} \multicolumn{1}{l|}{MixupMP (r=1.0, $\alpha$=0.1)} &$81.64$ &$0.6174$ &$3.99$ &$57.82$ &$1.8428$ &$1.60$\\ 
\rowcolor[HTML]{E6ECFC} \multicolumn{1}{l|}{MixupMP (r=0.1, $\alpha$=0.4)} &$80.53$ &$0.6512$ &$4.70$ &$56.86$ &$1.9375$ &$3.94$\\ 
\rowcolor[HTML]{E6ECFC} \multicolumn{1}{l|}{MixupMP (r=1.0, $\alpha$=0.4)} &$83.46$ &$0.5558$ &$1.82$ &$60.39$ &$1.7103$ &$\mathbf{1.12}^*$\\ 
\rowcolor[HTML]{E6ECFC} \multicolumn{1}{l|}{MixupMP (r=0.1, $\alpha$=1.0)} &$81.24$ &$0.6281$ &$4.76$ &$58.17$ &$1.8669$ &$2.72$\\ 
\rowcolor[HTML]{E6ECFC} \multicolumn{1}{l|}{MixupMP (r=1.0, $\alpha$=1.0)} &$83.90$ &$0.5371$ &$1.61$ &$61.82$ &$1.6101$ &$1.82$\\ 
\rowcolor[HTML]{E6ECFC} \multicolumn{1}{l|}{MixupMP (r=0.1, $\alpha$=2.0)} &$82.02$ &$0.6036$ &$4.12$ &$58.30$ &$1.8550$ &$3.28$\\ 
\rowcolor[HTML]{E6ECFC} \multicolumn{1}{l|}{MixupMP (r=1.0, $\alpha$=2.0)} &$\mathbf{85.12}^*$ &$\mathbf{0.4840}^*$ &$\mathbf{1.03}^*$ &$\mathbf{62.43}^*$ &$\mathbf{1.5632}^*$ &$2.43$\\ 
\multicolumn{1}{l|}{Mixup Ensemble ($\alpha$=0.1)} &$82.80$ &$0.5649$ &$1.30$ &$57.65$ &$1.8422$ &$1.29$\\ 
\multicolumn{1}{l|}{Mixup Ensemble ($\alpha$=0.4)} &$83.91$ &$0.5468$ &$3.27$ &$60.56$ &$1.6720$ &$2.81$\\ 
\multicolumn{1}{l|}{Mixup Ensemble ($\alpha$=1.0)} &$84.05$ &$0.5378$ &$5.43$ &$61.44$ &$1.6322$ &$7.17$\\ 
\multicolumn{1}{l|}{Mixup Ensemble ($\alpha$=2.0)} &$82.94$ &$0.5886$ &$9.30$ &$62.18$ &$1.6021$ &$9.37$\\ 
\hline 
\multicolumn{1}{l|}{\textbf{Implicit ensemble (B=20)}}  &  &  &  &  &  & \\ 
\multicolumn{1}{l|}{MC Dropout} &$\mathbf{82.87}$ &$0.6778$ &$8.72$ &$50.58$ &$2.4267$ &$15.68$\\ 
\rowcolor[HTML]{E6ECFC} \multicolumn{1}{l|}{MixupMP-MC (r=0.1, $\alpha$=0.1)} &$77.17$ &$0.8627$ &$10.66$ &$52.73$ &$2.2254$ &$10.52$\\ 
\rowcolor[HTML]{E6ECFC} \multicolumn{1}{l|}{MixupMP-MC (r=1.0, $\alpha$=0.1)} &$79.55$ &$0.7213$ &$7.32$ &$54.27$ &$2.1069$ &$8.31$\\ 
\rowcolor[HTML]{E6ECFC} \multicolumn{1}{l|}{MixupMP-MC (r=0.1, $\alpha$=0.4)} &$78.21$ &$0.7938$ &$8.94$ &$53.42$ &$2.1588$ &$8.59$\\ 
\rowcolor[HTML]{E6ECFC} \multicolumn{1}{l|}{MixupMP-MC (r=1.0, $\alpha$=0.4)} &$81.49$ &$0.6429$ &$4.94$ &$56.24$ &$1.9447$ &$4.81$\\ 
\rowcolor[HTML]{E6ECFC} \multicolumn{1}{l|}{MixupMP-MC (r=0.1, $\alpha$=1.0)} &$79.28$ &$0.7276$ &$7.47$ &$54.30$ &$2.1346$ &$8.80$\\ 
\rowcolor[HTML]{E6ECFC} \multicolumn{1}{l|}{MixupMP-MC (r=1.0, $\alpha$=1.0)} &$81.56$ &$0.6472$ &$4.60$ &$57.40$ &$1.8638$ &$4.72$\\ 
\rowcolor[HTML]{E6ECFC} \multicolumn{1}{l|}{MixupMP-MC (r=0.1, $\alpha$=2.0)} &$79.60$ &$0.7324$ &$7.58$ &$54.71$ &$2.0705$ &$7.69$\\ 
\rowcolor[HTML]{E6ECFC} \multicolumn{1}{l|}{MixupMP-MC (r=1.0, $\alpha$=2.0)} &$81.58$ &$0.6619$ &$5.40$ &$\mathbf{57.82}$ &$\mathbf{1.8104}$ &$3.27$\\ 
\multicolumn{1}{l|}{Mixup-MC ($\alpha$=0.1)} &$79.50$ &$0.7097$ &$5.71$ &$53.65$ &$2.1018$ &$7.61$\\ 
\multicolumn{1}{l|}{Mixup-MC ($\alpha$=0.4)} &$81.49$ &$\mathbf{0.6356}$ &$\mathbf{2.37}$ &$55.75$ &$1.9766$ &$3.09$\\ 
\multicolumn{1}{l|}{Mixup-MC ($\alpha$=1.0)} &$80.86$ &$0.6581$ &$2.98$ &$56.54$ &$1.8723$ &$\mathbf{2.33}$\\ 
\multicolumn{1}{l|}{Mixup-MC ($\alpha$=2.0)} &$80.27$ &$0.6964$ &$6.19$ &$56.47$ &$1.8432$ &$2.77$\\ 
\multicolumn{1}{l|}{CAMixup-MC ($\alpha$=0.1)} &$78.33$ &$0.7711$ &$6.22$ &$52.86$ &$2.1693$ &$10.39$\\ 
\multicolumn{1}{l|}{CAMixup-MC ($\alpha$=0.4)} &$77.72$ &$0.8392$ &$7.44$ &$53.07$ &$2.1843$ &$11.49$\\ 
\multicolumn{1}{l|}{CAMixup-MC ($\alpha$=1.0)} &$78.68$ &$0.7890$ &$7.11$ &$52.28$ &$2.2467$ &$11.96$\\ 
\multicolumn{1}{l|}{CAMixup-MC ($\alpha$=2.0)} &$79.52$ &$0.7728$ &$8.49$ &$52.68$ &$2.2535$ &$11.94$\\ 
\hline 
 \end{tabular}